\documentclass[10pt,twocolumn,letterpaper]{article}

\usepackage{wacv}              

\usepackage{graphicx}
\usepackage{amsmath}
\usepackage{amssymb}
\usepackage{nicematrix}
\usepackage{booktabs}
\usepackage{threeparttable}  

\usepackage[pagebackref,breaklinks,colorlinks]{hyperref}
\usepackage{mathtools}
\usepackage{amssymb}
\usepackage{graphicx}
\usepackage{algpseudocode}
\usepackage{caption}
\usepackage{subcaption}
\usepackage{multirow}
\usepackage{tablefootnote}
\usepackage{wrapfig}
\usepackage{textcomp}
\usepackage{siunitx}
\usepackage{multirow}
\usepackage{wrapfig}
\usepackage{booktabs}
\usepackage{graphics}
\usepackage{multibib}
\usepackage{algorithm}
\usepackage{arydshln}

\usepackage[capitalize]{cleveref}
\crefname{section}{Sec.}{Secs.}
\Crefname{section}{Section}{Sections}
\Crefname{table}{Table}{Tables}
\crefname{table}{Tab.}{Tabs.}

\usepackage[accsupp]{axessibility}

\def\wacvPaperID{2215} 
\def\confName{WACV}
\def\confYear{2024}

\begin{document}

\title{Concept-Centric Transformers:\\Enhancing Model Interpretability through Object-Centric Concept Learning within a Shared Global Workspace}

\author{
Jinyung Hong\textsuperscript{1}, Keun Hee Park\textsuperscript{1}, and Theodore P.~Pavlic\textsuperscript{1, 2} \\
\textsuperscript{1}School of Computing and Augmented Intelligence \\
    \textsuperscript{2}School of Life Sciences\\
Arizona State University\\
Tempe, AZ 85281\\
{\tt\small \{ jhong53, kpark53, tpavlic \}@asu.edu}
}
\maketitle

\begin{abstract}
Many interpretable AI approaches have been proposed to provide plausible explanations for a model's decision-making. However, configuring an explainable model that effectively communicates among computational modules has received less attention. A recently proposed shared global workspace theory showed that networks of distributed modules can benefit from sharing information with a bottlenecked memory because the communication constraints encourage specialization, compositionality, and synchronization among the modules. Inspired by this, 
we propose Concept-Centric Transformers, a simple yet effective configuration of the shared global workspace for interpretability, consisting of: i)~an object-centric-based memory module for extracting semantic concepts from input features, ii)~a cross-attention mechanism between the learned concept and input embeddings, and iii)~standard classification and explanation losses to allow human analysts to directly assess an explanation for the model's classification reasoning. 
We test our approach against other existing concept-based methods on classification tasks for various datasets, including CIFAR100, CUB-200-2011, and ImageNet, and we show that our model achieves better classification accuracy than all baselines across all problems but also generates more consistent concept-based explanations of classification output.
\end{abstract}

\section{Introduction}
\label{sec:introduction}

Although state-of-the-art machine-learning models have achieved remarkable performance across a wide range of applications, their intrinsic lack of transparency due to their many degrees of training freedom limits their usage in safety-critical areas---such as medical diagnostics, healthcare, public infrastructure safety, and visual inspection for civil engineering---where trustworthy domain-specific knowledge is crucial for decision making. Recently, several developed methods provide \emph{post hoc} explanations that identify relevant features that a trained model uses to make predictions~\cite{ribeiro2016should, selvaraju2017grad, lundberg2017unified, smilkov2017smoothgrad}, but these are commonly criticized for focusing only on low-level features~\cite{kim2018interpretability, alvarez2018towards, kindermans2019reliability, su2019one}. In contrast, \emph{intrinsically interpretable models}~\cite{rudin2019stop} have been proposed to make decisions based on human-understandable ``concepts,'' the foundation of domain expertise~\cite{barbiero2022entropy, ghorbani2019towards, kim2018interpretability, yeh2020completeness, koh2020concept, zarlenga2022concept, goyal2019explaining, kazhdan2020now, chen2020concept, alvarez2018towards, li2018deep, chen2019looks, rigotti2021attention}.
The gap between \emph{post hoc} explainability and intrinsically interpretable models is also discussed in the NLP community concerning interpreting \emph{attention mechanisms}~\cite{bahdanau2014neural}. In particular, the debate over what degree of interpretability can be ascribed to attention weights over input tokens still needs to be settled to help meet the need for interpretability of attention mechanisms~\cite{jain2019attention, wiegreffe2019attention, serrano2019attention, alvarez2018towards, carvalho2019machine}.

Ideally, an intrinsically interpretable model will generate explanations that are compositions of individually meaningful modules. Modular explanations may improve the human understanding, and natural neuronal systems that continue to inspire AI development are often described as having modular architectures themselves~\cite{robbins2017modularity, ballard1986cortical, baldwin2000design, brooks1991intelligence}. Thus, structuring algorithms to promote learning of modular latent structures may also lead to better overall performance. Motivated by improving modularity in interpretable models, we propose the \emph{Concept-Centric Transformer~(CCT)}, a framework of intrinsically interpretable models inspired by the Shared Global Workspace~(SGW)~\cite{goyal2021coordination}, a new conceptual framework meant to generally encourage modularity by forcing parallel specialized components to compete for bottlenecked access to a shared memory. The configuration of CCT allows trained models to have simple, modular structures that can extract semantic concepts with or without the guidance of ground-truth explanations of the concepts.

In what follows, we frame our CCT as a novel extension of the SGW concept to interpretable model development, and we describe how CCT is implemented using three key components: i)~\textbf{Concept-Slot-Attention~(CSA) module} that interfaces with image embedding from a backbone model and produces a set of task-dependent embeddings for concepts, ii)~\textbf{Cross-Attention~(CA) module} that generates classification outputs using cross-attention between input features and the CSA module's embeddings, and iii)~specialized loss penalties, including \textbf{Explanation Loss}, when expert's knowledge can be leveraged, and \textbf{Sparsity Loss}, an entropy-based loss to enforce the sparsity to determine the importance of features during training. The CCT architecture is designed to augment existing deep-learning backbones to add explainability to them. Consequently, we validate our approach on three image benchmark datasets---CIFAR100 Super-class~\cite{fischer2019dl2}, CUB-200-2011~\cite{wah2011caltech}, and ImageNet~\cite{deng2009imagenet}---combining it with various deep-learning backbones, such as Vision Transformer~(ViT)~\cite{dosovitskiy2020image}, Swin Transformer~(SwinT)~\cite{liu2021swin}, and ConvNeXt~\cite{liu2022convnet}. 

\section{Related Work}
\label{sec:related_work}
Significant advances have recently been made in devising explainable and interpretable models to measure the importance of individual features for predictive output. The \emph{post hoc} analysis is one general approach to analyzing a trained model by matching explanations to classification outputs~\cite{alvarez2018towards, ribeiro2016should, lundberg2017unified}. For example, activation maximization~\cite{van2016pixel, nguyen2016synthesizing, yosinski2015understanding} and saliency visualization~\cite{selvaraju2017grad, smilkov2017smoothgrad, sundararajan2017axiomatic} are well-known methods for CNNs. 
Attention-based interpretable approaches have also been introduced to identify the most relevant parts of the input that the network focuses on when making a decision~\cite{zhang2014part, zhou2016learning, zhou2018interpretable, zheng2017learning, fu2017look, girshick2014rich, girshick2015fast, huang2016part}. 

In addition, designing methods that explain predictions with high-level, human-understandable concepts~\cite{ghorbani2019towards, kim2018interpretability, yeh2020completeness, koh2020concept, zarlenga2022concept, goyal2019explaining, kazhdan2020now, chen2020concept, barbiero2022entropy, li2018deep, xue2022protopformer, chen2019looks, rigotti2021attention} is one of the recent advancements in the field of interpretability. These intrinsically interpretable methods focus on identifying common activation patterns in the nodes of the last layer of the neural network corresponding to human-understandable categories or constraining the network to learn such concepts. Among them, our work is most similar to \emph{Concept Transformers~(CTs)}~\cite{rigotti2021attention}, a framework that learns high-level concepts defined with a set of related dimensions. Those concepts, which can be part-specific or global, typically can boost the performance of the learning task while offering explainability at no additional cost to the network. 
However, that approach relies on extracting concepts based on provided image patches even though each image patch may be an unreliable predictor of high-level concepts. Our CCT formulation generalizes this approach beyond image patches.

Historically, it has been argued that it is better to build an intelligent system from many interacting specialized modules rather than a single ``monolithic'' entity to deal with a broad spectrum of conditions and tasks~\cite{minsky1988society, robbins2017modularity, goyal2022inductive}. Thus, there has been significant effort on synchronization between computationally specialized modules via a shared global workspace~\cite{dehaene2011experimental, santoro2018relational, goyal2019recurrent, munkhdalai2019metalearned, jaegle2021perceiver}. Furthermore, work on the integration of modular computational architectures with working memories takes inspiration from biology, neuroscience, and cognitive science~\cite{ramsauer2020hopfield, hong2021insect, hong2022representing, wu2022memvit}. The recently proposed shared global workspace~\cite{goyal2021coordination} shows how to utilize the attention mechanism to encourage the most helpful information to be shared among neural modules in modern AI frameworks. This approach is the inspiration for our use of an explicit working memory in the CCT to improve the generalization of Transformer- and object-centric-based models in the context of explainable models.

\section{Preliminary}
\label{sec:pre}

\begin{figure}[t!]
    \centering
    \includegraphics[width=0.4\textwidth]{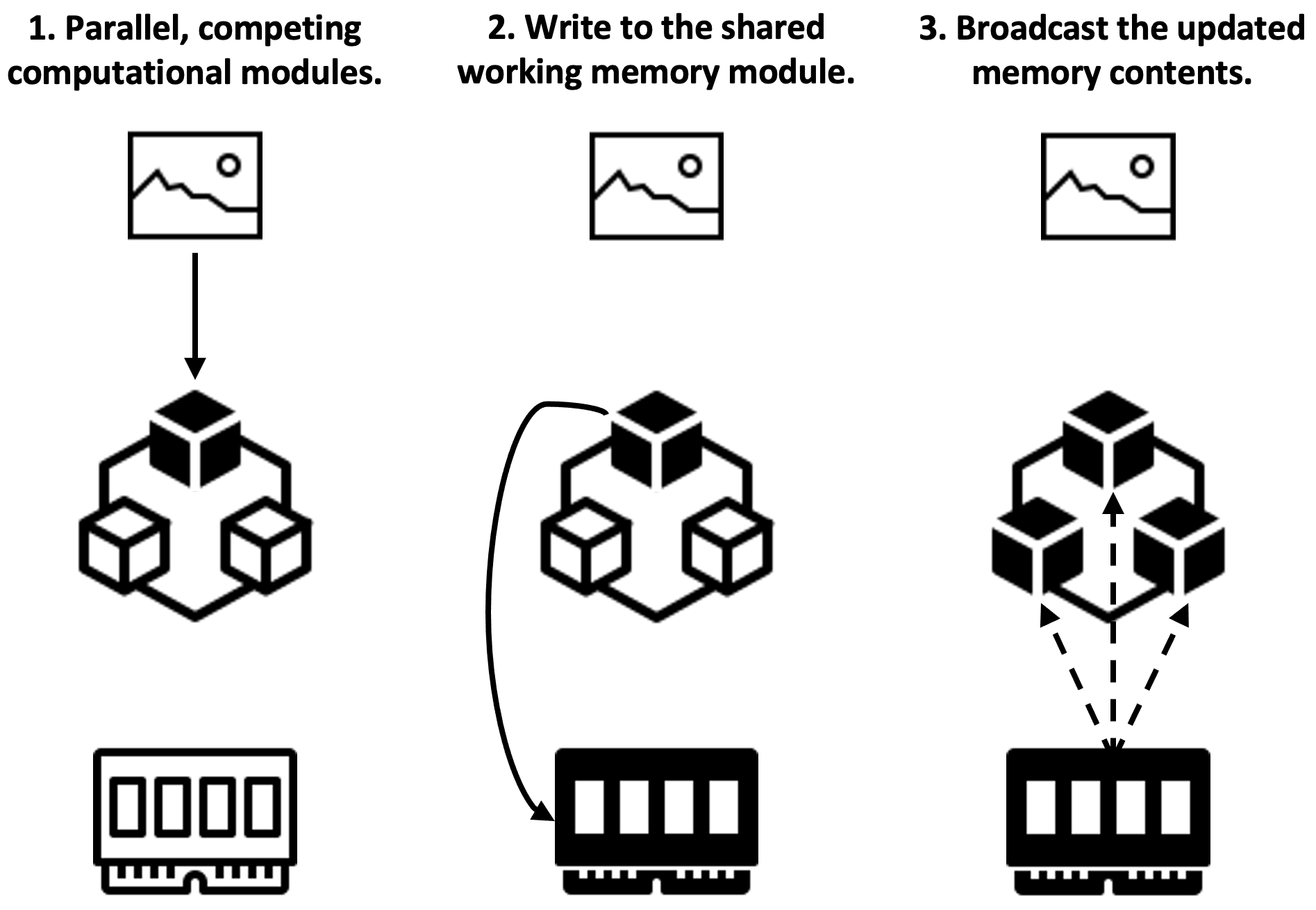}
    \caption{The shared global workspace~\cite{goyal2021coordination} emerges from three steps: 1)~A collection of computational modules (or \emph{specialists}) perform standard processing, and a subset of the specialists becomes active at a particular computational stage depending upon the input; 2)~The active specialist writes information in a shared working memory module (or \emph{shared workspace}); 3)~The updated contents of the workspace are broadcast to all specialists. We explore how these steps can be used to add explainability in AI frameworks. The generic figure above is inspired by~\protect\cite[Fig.~1]{goyal2021coordination}.}
    \label{fig:shared_workspace}
\end{figure}

\paragraph{The Shared Global Workspace in AI Models.}
Inspired by the Global Workspace Theory~(GWT) from cognitive science~\cite{baars1993cognitive, dehaene1998neuronal, shanahan2005applying, shanahan2006cognitive, shanahan2010embodiment, shanahan2012brain, dehaene2021consciousness}, the Shared Global Workspace~(SGW)~\cite{goyal2021coordination} 
explores how GWT can be manifested in modern AI models to possess communication and coordination schemes where several sparsely communicating \emph{specialists}~(specific computing modules dealing with the input) interact via a \emph{shared workspace}~(a shared working memory module). To do so, the transformer and slot-based methods were extended by adding a shared workspace and allowing the modules to compete for write access in each computational stage~(Fig.~\ref{fig:shared_workspace}). Replacing pairwise communications among the modules with interaction facilitated by the shared workspace allows for: i)~higher-order interaction among the modules, ii)~dynamic filtering due to the memory persistence, and iii)~computational sophistication of using shared workspace for synchronizing different specialists.
Motivated by the SGW, we aim to discover an efficient configuration for intrinsically interpretable AI frameworks and propose a simple but efficient way of interacting between a shared working space and specialists to improve interpretability and performance. 

\paragraph{Variants of Slot-Attention.}
Due to its simple yet effective design, Slot-Attention~(SA)~\cite{locatello2020object} has gained significant attention in unsupervised object-centric learning to mimic the development of symbolic understanding in human cognition.
The iterative attention mechanism allows SA to learn and compete between slots for explaining parts of the input, showing a soft clustering effect on visual inputs~\cite{locatello2020object}. However, as revealed by recent studies, the vanilla SA module as innately limited in that: i)~the random initialization for slots hampers addressing object-binding in input and ii)~it heavily depends on hyperparameter tuning so that it cannot generally be applicable in many domains. Thus, some variants of SA, including I-SA~\cite{chang2022object} and BO-QSA~\cite{jia2022improving}, have been proposed recently to address those issues\footnote{We omit details of their technical differences as they are out of scope.}.

There are several existing examples on leveraging slot-based methods in explainable models to extract semantic concepts~\cite{li2021scouter, wang2023learning}. However, few studies have emphasized the perspective of modular architecture to foster communication between the slot-based model and other modules. We leverage the three SA variants above as the shared workspace of the SGW and explore how to encourage the interactions among them to achieve better interpretability and performance.

\section{Concept-Centric Transformers}
\label{sec:method}

\begin{figure*}[t!]
     \centering
     \begin{subfigure}[b]{0.575\textwidth}
         \centering
         \includegraphics[width=\textwidth]{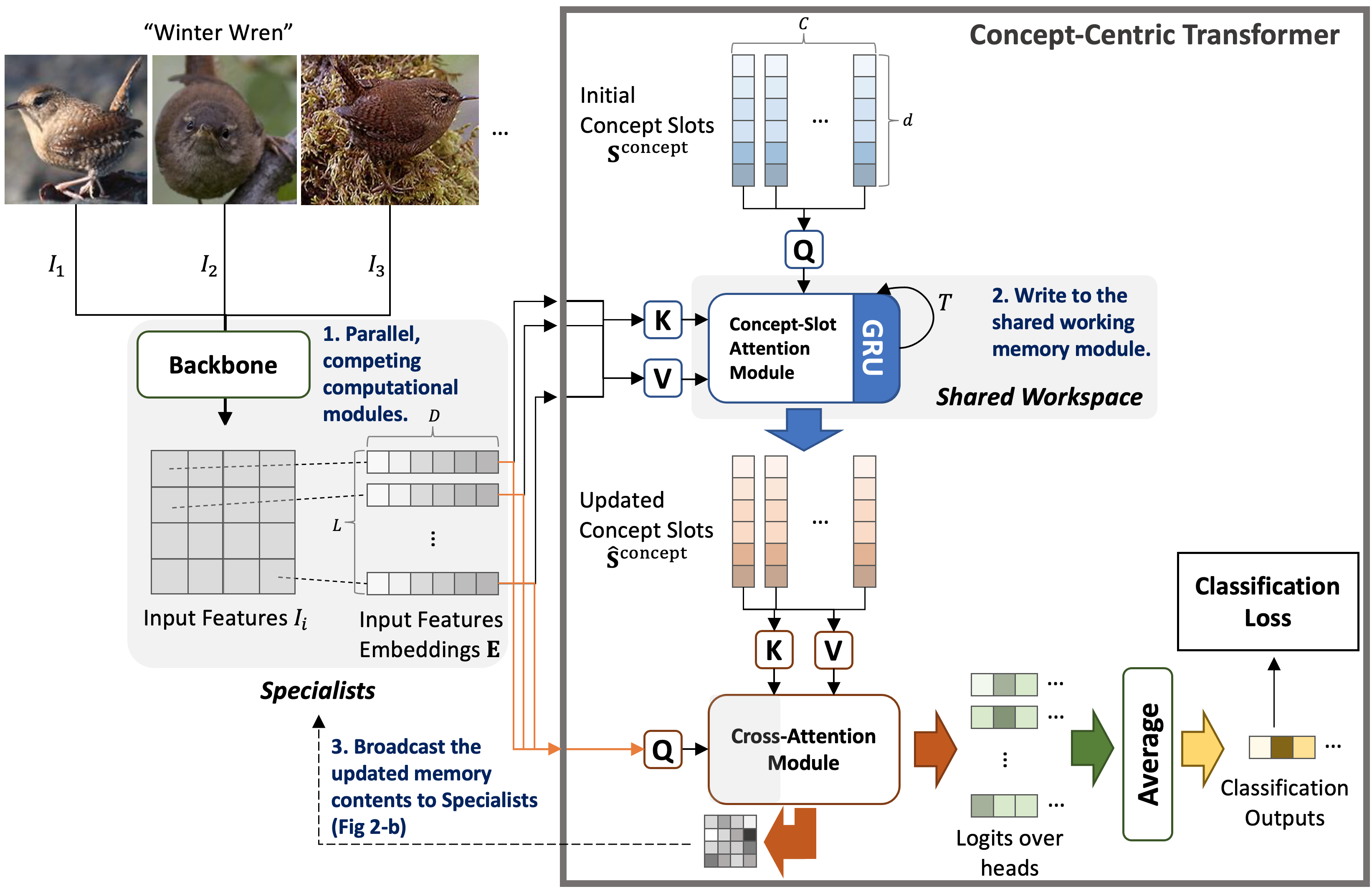}
         \caption{Concept-Centric Transformers via a Shared Workspace}
         \label{fig:main_architecture}
     \end{subfigure}
     \hspace{20pt}
     \begin{subfigure}[b]{0.32\textwidth}
         \centering
         \includegraphics[trim={0pt 0pt 5pt 0pt},clip,width=0.75\textwidth]{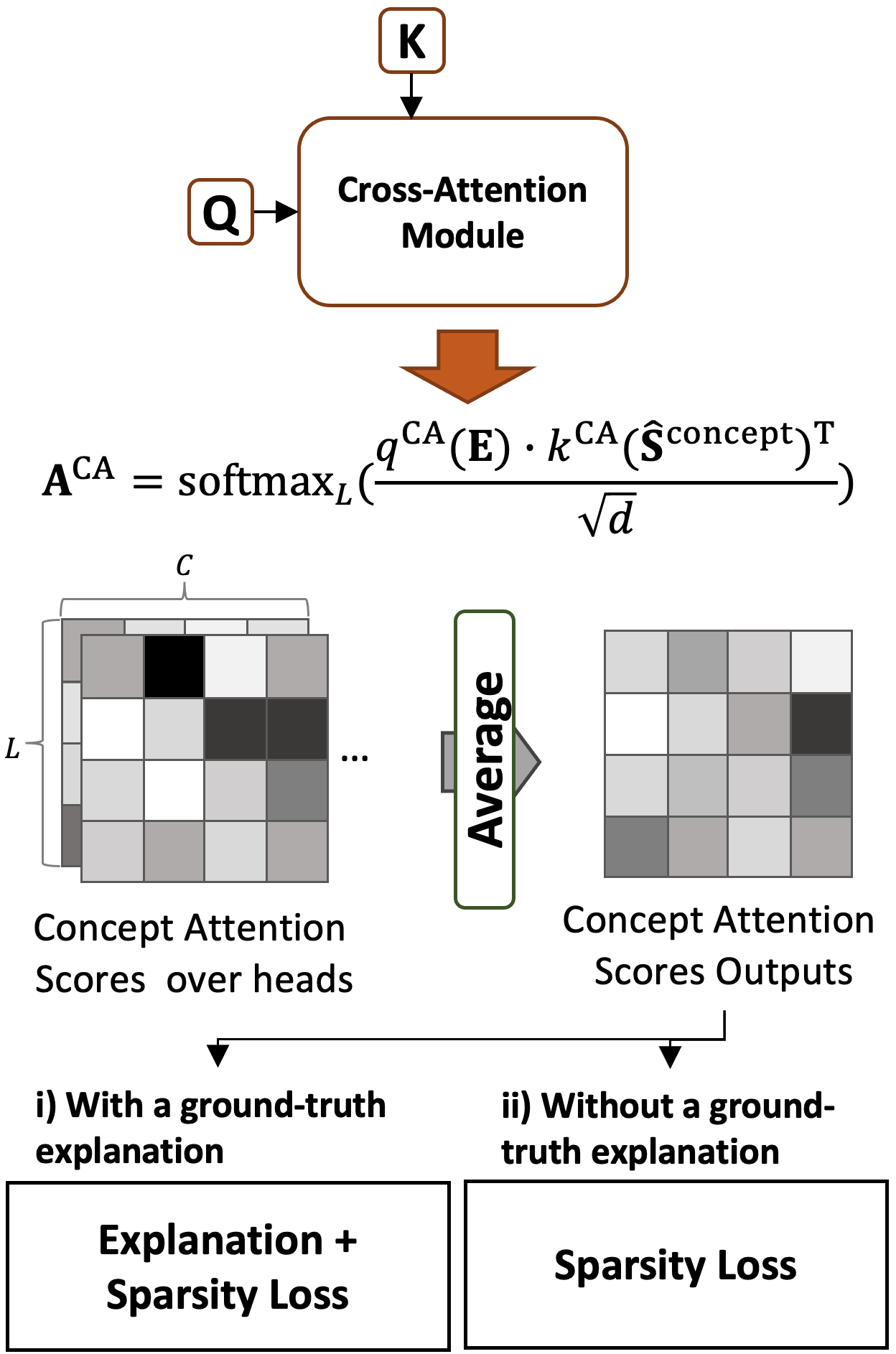}
         \caption{Interpretable Broadcast to Specialists}
         \label{fig:interpretable_broadcast}
     \end{subfigure}
    \caption{Overall architecture of Concept-Centric Transformers~(CCTs). \subref{fig:main_architecture} The CCT~(in a solid gray line) is a drop-in replacement for the classifier head of any backbone architecture, such as a ViT or a CNN, and consists of two modules: (1)~Concept-Slot-Attention module and (2)~Cross-Attention module.  \subref{fig:interpretable_broadcast} The process of training uses losses to induce an interpretable broadcast scheme.}
    \label{fig:overall_architecture}
\end{figure*}

For supervised classification tasks, we introduce Concept-Centric Transformers~(CCTs), an instantiation of the SGW for configuring an intrinsically interpretable model, and we will describe the connection between the SGW steps~(Fig.~\ref{fig:shared_workspace}) and our formulation. Our model consists of: i)~\emph{Concept-Slot-Attention~(CSA)} module that acts as a shared memory module and extracts the latent concept embedding specific to each batch of input, ii)~\emph{Cross-Attention~(CA)} module for broadcasting between input embedding and the extracted concept embedding from the CSA module so that it produces classification output as well as faithful and plausible concept-based explanations and encourages pairwise interactions among them, and iii)~specialized losses, including \emph{Explanation Loss} and \emph{Sparsity Loss}, which is our information broadcast scheme to encourage interpretability. The CCT architecture, summarized in Fig.~\ref{fig:overall_architecture}, is described in the following sections.
%
Further details, including limitations (Appendix~\ref{app:limit}), are given in the appendix.

\subsection{Between Specialists and the Shared Workspace}
\label{subsec:step1_step2}

Following the general structure of the SGW, we leverage \emph{specialists}, which are the computational modules of our backbone, and a \emph{shared workspace} by utilizing slot-based methods in our CSA module, including SA~\cite{locatello2020object}, I-SA~\cite{chang2022object}, and BO-QSA~\cite{jia2022improving}. Because of the modularity in our formulation, those three SA variants are interchangeable, and we demonstrate the performance comparison among them in our experiments. We use a conventional key--query--value attention mechanism to implement the competition between specialists to write into the workspace, similar to the SGW. The CSA module encodes a set of $L$ input feature vectors $\mathbf{E}$ into concept representations $\mathbf{S}^{\mathtt{concept}}$, which we refer to as \emph{concept slots}. 


\paragraph{Concept Binding Specific to Each Input Batch.}
With the number of concepts $C$, the concept slots $\mathbf{S}^{\mathtt{concept}} \in \mathbb{R}^{C \times d}$ first perform competitive attention~\cite{locatello2020object} on the input features $\mathbf{E} \in \mathbb{R}^{L \times D}$. For this, we apply linear projection $q^{\mathtt{CSA}}$ on the concept slots to obtain the queries and projections $k^{\mathtt{CSA}}$ and $v^{\mathtt{CSA}}$ on the inputs to obtain the keys and the values, all having the same size $d$~\footnote{For simplicity, embedding size $d$ is shared equally in our method.}. Then, we perform a dot product between the queries and keys to get the attention matrix $\mathbf{A}^{\mathtt{CSA}} \in \mathbb{R}^{C \times L}$. In $\mathbf{A}^{\mathtt{CSA}}$, each entry $\mathbf{A}_{c,l}^{\mathtt{CSA}}$ is the attention weight of concept slot $c$ for attending over the input vector $l$. We normalize $\mathbf{A}^{\mathtt{CSA}}$ by applying softmax across concept slots, i.e., along the axis $C$. This implements a form of competition among slots for attending to each input $l$.

We then seek to group and aggregate the attended inputs and obtain the attention readout for each concept slot. Intuitively, this represents how much the attended inputs contribute to semantically representing each concept. For this, we normalize the attention matrix $\mathbf{A}^{\mathtt{CSA}} \in \mathbb{R}^{C \times L}$ along the axis $L$ and multiply it with the input values $v^{\mathtt{CSA}}(\mathbf{E}) \in \mathbb{R}^{L \times d}$. This produces the attention readout in the form of a matrix $\mathbf{U} \in \mathbb{R}^{C \times d}$ where each row $u_{c} \in \mathbb{R}^{d}$ is the readout corresponding to concept slot $c$; $\mathbf{A}^{\mathtt{CSA}} = \operatorname{softmax}_{C}(q^{\mathtt{CSA}}(\mathbf{S}^{\mathtt{concept}}) \cdot k^{\mathtt{CSA}}(\mathbf{E})^{\top}/ \sqrt{d})$,
$\mathbf{A}_{c,l}^{\mathtt{CSA}} = \mathbf{A}_{c,l}^{\mathtt{CSA}}/ \sum_{l=1}^{L} \mathbf{A}_{c,l}^{\mathtt{CSA}}$
, and $\mathbf{U} = \mathbf{A}^{\mathtt{CSA}} \cdot v^{\mathtt{CSA}}(\mathbf{E})$.
We use the readout information obtained from concept binding and update each concept slot. The aggregated updates $\mathbf{U}$ are finally used to update the concept slots via a learned recurrent function, for which we use a Gated Recurrent Unit~(GRU)~\cite{cho2014learning} with $d$ hidden units so that $\mathbf{S}^{\mathtt{concept}} = \mathtt{GRU}(\mathbf{S}^{\mathtt{concept}}, \mathbf{U})$.
The processes above form one refinement iteration. The concept slots obtained from the last iteration are considered final. The overall module is described in Algorithm~\ref{alg:csa_module} in pseudo-code in Appendix~\ref{app:method_detail}.

\subsection{Broadcast Updated Memories to Specialists}
\label{subsec:broadcast}

At this stage of the SGW, each specialist must update its status using information broadcast from the shared workspace. We also leverage the cross-attention mechanism (called the CA module) to make specialists queries (step 3 in Figs~\ref{fig:shared_workspace} and~\ref{fig:overall_architecture}) and perform dot products between them and the values from the updated concept slots to update the state of each specialist. However, because we aim to configure an interpretable model and perform classification tasks, we modify the iterative process by combining it with our desired downstream classification task: i) guided by expert knowledge if ground-truth concept explanations are available, or ii) using only sparsity loss to enforce minimizing the entropy of the broadcasting information.

A set of $L$ input feature vectors $\mathbf{E} \in \mathbb{R}^{L \times D}$ are re-used with a linear projection $q^{\mathtt{CA}}$ to attain the queries, and projections $k^{\mathtt{CA}}$ and $v^{\mathtt{CA}}$ are applied to the extracted concept slots with position embedding $\hat{\mathbf{S}}$ from the CSA module. The resulting keys and values are used in a cross-attention mechanism with the queries, and the cross-attention then outputs an attention weight $\mathbf{A}^{\mathtt{CA}} = \operatorname{softmax}_{L}\left(q^{\mathtt{CA}}(\mathbf{E}) \cdot k^{\mathtt{CA}}(\hat{\mathbf{S}}^{\mathtt{concept}})^{\top} / \sqrt{d} \right) \in \mathbb{R}^{L \times C}$
between each patch--concept slot pair. The final output of the CA module is the product obtained by multiplying the attention map $\mathbf{A}^{\mathtt{CA}}$, the values $v^{\mathtt{CA}}(\hat{\mathbf{S}}^{\mathtt{concept}}) \in \mathbb{R}^{C \times d}$, and an output matrix $\mathbf{O} \in \mathbb{R}^{d \times n_{c}}$ that projects onto the (unnormalized) $n_{c}$ logits over the output classes and then averaging over input features\footnote{For simplicity, we describe a single-head attention model here; a multi-head version~\protect\cite{vaswani2017attention} is available and is also used in our experiments.}; that is, for $i=1,\dots,n_{c}$,
\begin{equation}
    \text{logit}_{i} = \frac{1}{L}\textstyle\sum_{l=1}^{L}\mathbf{A}_{l}^{\mathtt{CA}} \cdot v^{\mathtt{CA}}(\hat{\mathbf{S}}^{\mathtt{concept}}) \cdot \mathbf{O}_{:,i}
    \label{eq:logit}
\end{equation}
So, given an input $\mathbf{x}$ to the network, the conditional probability of output class $i \in \{1,\dots,n_{c}\}$ is:
\begin{equation}
    \text{Pr}(i|\mathbf{x}) = \operatorname{softmax}_{i}\left(\textstyle\sum_{c=1}^{C}\beta_{c}\gamma_{c}(\mathbf{x})\right) \quad 
    \label{eq:ca_prob}
\end{equation}
with $\beta_{c}$ components $\beta_{ci} \coloneqq (v^{\mathtt{CA}}(\hat{\mathbf{S}}^{\mathtt{concept}}) \cdot \mathbf{O})_{c,i}$
where $\gamma_{c}(\mathbf{x})$ are non-negative relevance scores that depend on $\mathbf{x}$ through the averaged attention weights; that is, $\gamma_{c}(\mathbf{x}) = (1/L) \sum_{l=1}^{L}\mathbf{A}_{l,c}^{\mathtt{CA}}$.
We can interpret the equations above from the two following perspectives:

\paragraph{1) Faithful Concept-slot-based Explanations by Design.} 
The CA module output is a multinomial logistic regression model over positive variables $\gamma_{c}(\mathbf{x})$ that measures the contribution of each concept slot. \emph{Faithfulness} is the degree to which explanation reflects the decision and aims to ensure that the explanations are indeed explaining model operation~\cite{lakkaraju2019faithful, guidotti2018survey}. As shown in~\cite{rigotti2021attention}, the result of the CA module follows the linear relation between the value vectors and the classification logits and comes from the design choices of computing outputs from the value matrix $v^{\mathtt{CA}}(\hat{\mathbf{S}}^{\mathtt{concept}})$ through the linear projection $v^{\mathtt{CA}}(\hat{\mathbf{S}}^{\mathtt{concept}}) \cdot \mathbf{O}$ and aggregating patch contributions by averaging. So, our CA module is also guaranteed to be faithful by design by satisfying Proposition~1 in~\cite{rigotti2021attention} and the technical definitions of \emph{faithfulness} from~\cite{alvarez2018towards}.

\paragraph{2) Dynamic State Update for Specialists with Information Broadcast.}

In the original definition of the SGW, $\mathbf{A}^{\mathtt{CA}} \cdot v^{\mathtt{CA}}(\hat{S}^{\mathtt{concept}})$ is the formal computation of the update for specialists~(step 3 from~\cite[Sec~2.1]{goyal2021coordination}). Instead of applying an additional iterative process of updating specialists, Eqs~\ref{eq:logit} and~\ref{eq:ca_prob} are to produce classification outputs using the weight $\mathbf{O}$. So, the attention mask $\mathbf{A}^{\mathtt{CA}}$ can contain not only information from the updated memory but also classification error. Furthermore, by directly manipulating the mask $\mathbf{A}^{\mathtt{CA}}$, we finally define explanation loss and sparsity loss to enhance the model's explainability. 

%

\subsection{Training Objectives for Interpretability}
\label{subsec:training}

\paragraph{Plausibility by Construction with Explanation Loss.}
\emph{Plausibility} refers to how convincing the interpretation is to humans~\cite{guidotti2018survey, carvalho2019machine}. To provide plausible human-understandable explanations, we leverage the idea of explicitly guiding the attention heads to focus on concepts in the input based on domain expertise that are important for correctly classifying the input.
%
Similar to~\cite{deshpande2020guiding}, given a desired distribution of attention $\mathbf{H}$ provided by domain knowledge, the attention weights from the CA module of the CCT $\mathbf{A}^{\mathtt{CA}}$ are used as a regularization term by adding an \emph{explanation cost} to the objective function that is proportional to $\mathcal{L}_{expl} = \lVert \mathbf{A} - \mathbf{H} \rVert^{2}_{F}$, where $\lVert \cdot \rVert$ is the Frobenius norm. The ground-truth explanation $\mathbf{H}$ can indicate global (e.g., sub-class or dominant attribute of a bird) or spatial/image-patch-level information (e.g., eye color of a bird). Below, we demonstrate the effectiveness of this loss in the experiments of CIFAR100~Super-class and CUB-200-2011.

\paragraph{Sparsity Loss based on Entropy.}
The advantage of richer, more informative labels can increase interpretability, but this comes at the expense of additional annotation effort. A methodology that can bypass this trade-off would be particularly worthwhile. We can attain this capability through our configuration that involves interactions between specialists and a shared workspace.
We introduce the sparsity loss based on minimizing the entropy of the attention mask $\mathbf{A}^{\mathtt{CA}}$; $\mathcal{L}_{sparse} = H(\mathbf{A}) = H(a_{1}, \dots, a_{|\mathbf{A}|}) = (1/|\mathbf{A}|) \sum_{i} -a_{i} \cdot \log(a_{i})$.
This loss can be used in the experiments with/without the ground-truth explanations.

\paragraph{Final Loss.}
Thus, the final loss to train the model becomes $\mathcal{L} = \mathcal{L}_{cls} + \lambda_{expl} \mathcal{L}_{expl} + \lambda_{sparse} \mathcal{L}_{sparse}$, where $\mathcal{L}_{cls}$ denotes the conventional classification loss. Notice that the constant $\lambda_{expl} \geq 0$ controls the relative contribution of the explanation loss to the total loss so our model can be applied with ground-truth explanations~($\lambda_{expl} > 0$) or without them~($\lambda_{expl} = 0$). Finally, the constant $\lambda_{sparse}$ handles the intensity of sparsity loss. Our experiments demonstrate that our model can perform well without using additional complicated losses, such as contrastive or reconstruct losses.


\section{Experiments}
\label{sec:experiments}
We evaluate the performance of our CCT on three distinct datasets:  CIFAR100 Super-class~\cite{fischer2019dl2}, CUB-200-2011~\cite{wah2011caltech}, and ImageNet~\cite{deng2009imagenet}. Specific objectives guide the selection of these datasets. Firstly, the CIFAR100 Super-class is a testing ground to demonstrate our model's exceptional capabilities under fully supervised conditions, where complete global concept explanations are available. Secondly, CUB-200-2011 acts as an intermediary dataset, allowing us to showcase our model's prowess in both supervised and unsupervised explanation setups. Lastly, we challenge our CCT with the ImageNet dataset, which represents a fully unsupervised scenario due to the absence of concept explanations. Despite this hurdle, we adapt our model to achieve remarkable performance even without any concept explanations. This comprehensive setup underscores our CCT's adaptability and versatility across various use cases, visual backbones, and data scenarios. Our experimental results are robustly validated through three different random seeds and $95\%$ confidence intervals, with additional details provided in Appendix~\ref{app:exp}.

\subsection{Evaluation on CIFAR100 Super-class}
\label{subsec:cifar100}
\begin{table}[t!]
    \centering
    \begin{NiceTabular}{l|c|c}
        Model &  F.C. Acc.~($\%$) & S.C. Acc.~($\%$) \\
        \midrule\midrule
        $\text{Vanilla ResNet}^{\dag}$ & NA & $83.2_{\pm 0.2}$ \\
        Vanilla ViT-T & NA & $86.2_{\pm 0.3}$ \\
        \midrule
        $\text{Hierarchical Model}^{\dag}$ & $71.2_{\pm 0.2}$ & $84.7_{\pm 0.1}$ \\  
        $\text{DL2}^{\dag}$~\cite{fischer2019dl2} & $75.3_{\pm 0.1}$ & $84.3_{\pm 0.1}$ \\ 
        $\text{MultiplexNet}^{\dag}$~\cite{hoernle2022multiplexnet} & $74.4_{\pm 0.2}$ & $85.4_{\pm 0.3}$ \\ 
        \midrule
        PIP-Net~\cite{Nauta_2023_CVPR} & NA & $83.9_{\pm 0.2}$ \\
        ProtoPFormer~\cite{xue2022protopformer} & NA & $81.7_{\pm 0.1}$\\
        ProtoPool~\cite{DBLP:journals/corr/abs-2112-02902} & NA & $82.9_{\pm 0.4}$ \\
        ProtoPNet~\cite{DBLP:journals/corr/abs-1806-10574} & NA & $82.3_{\pm 0.1}$\\
        Deform-ProtoPNet~\cite{DBLP:journals/corr/abs-2111-15000}  & NA & $83.7_{\pm 1.0}$ \\
        BotCL~\cite{wang2023learning} & NA & $56.9_{\pm 10}$ \\
        CT~\cite{rigotti2021attention} & $73.3_{\pm 2.9}$ & $92.1_{\pm 0.2}$ \\
        \midrule
        CCT: ViT-T+SA & $80.3_{\pm 0.4}$ & $92.6_{\pm 0.1}$ \\ 
        CCT: ViT-T+I-SA & $83.3_{\pm 0.1}$ & $92.8_{\pm 0.1}$  \\ 
        CCT: ViT-T+BO-QSA & $\textbf{83.4}_{\pm 0.1}$ & $\textbf{93.0}_{\pm 0.1}$ \\ 
    \bottomrule
    \end{NiceTabular}
    \caption{Test accuracy on fine-grained class~(F.C.) and super-class~(S.C.) label prediction on CIFAR100. Notice that the classification of super-classes and fine-grained classes are performed simultaneously and that this kind of experiment can be done in deep learning with constraints, but our method and CT are only among concept-based approaches. $\dag$ indicates results from~\cite{hoernle2022multiplexnet}.}
    \label{tab:eval_cifar100}
\end{table}
The CIFAR100 Super-class dataset is a variant of the CIFAR100~\cite{krizhevsky2009learning} image dataset.
It consists of 100 fine-grained classes~(F.C.) of images that are further grouped into 20 super-classes~(S.C.). For instance, the five fine-grained classes \emph{baby, boy, girl, man,} and \emph{woman} belong to the super-class \emph{people}. Since the introduction of \emph{deep-learning models with logical constraints}~\cite{fischer2019dl2}, this dataset has been used as one of the benchmark datasets to assess the effectiveness of embedding the constraints into neural networks~(in-depth surveys can be found in~\cite{dash2022review, giunchiglia2022deep}).
Our study highlights that the concepts within our CCT (and closely related CT~\cite{rigotti2021attention}) can be understood as \emph{constraints}, with differences outlined in Appendix~\ref{app_sub:cifar100}. We leverage individual fine-grained image classes as global concept explanations for a multi-class prediction task with 20 super-classes, employing a Vision Transformer-Tiny~(ViT-T)\footnote{For this experiment, Swin Transformer and ConvNeXt were not used as a backbone for our model because the number of parameters of two models (both SwinT-T and ConvNeXt-T are 28M) is larger than one of ViT-T.} as the backbone\footnote{For global concepts, we use the embedding of the CLS token as inputs.}.

Following~\cite{hoernle2022multiplexnet}, our CCT's performance is compared against three baseline groups. The first group includes vanilla backbone models like \emph{Wide ResNet 28-10}~\cite{zagoruyko2016wide} and \emph{ViT-T.} The second group involves neural-network models with logical constraints, including \emph{Hierarchical Model}~\cite{hoernle2022multiplexnet}, \emph{DL2}~\cite{fischer2019dl2}, and \emph{MultiplexNet}~\cite{hoernle2022multiplexnet}. The third group comprises concept-based explainable models; except for CT, these models cannot handle both fine-grained and super-class tasks concurrently. For our CCT's CSA module, we configure three SA variants---SA, I-SA, and BO-QSA---with the backbone and evaluate fine-grained class accuracy by comparing ground-truth concept explanations with top-1 predicted concepts based on the resulting attention scores.

%
Table~\ref{tab:eval_cifar100} presents the experimental outcomes, showing our CCT's substantial outperformance of all baselines. It significantly enhances ViT-T's backbone performance and achieves a remarkable increase in fine-grained class accuracy compared to CT. Notably, CT performs less effectively than logic-constraint-based approaches, highlighting our module's superior role in shaping global concepts.
%

\begin{figure}[t!]
    \centering
    \includegraphics[width=0.49\textwidth]{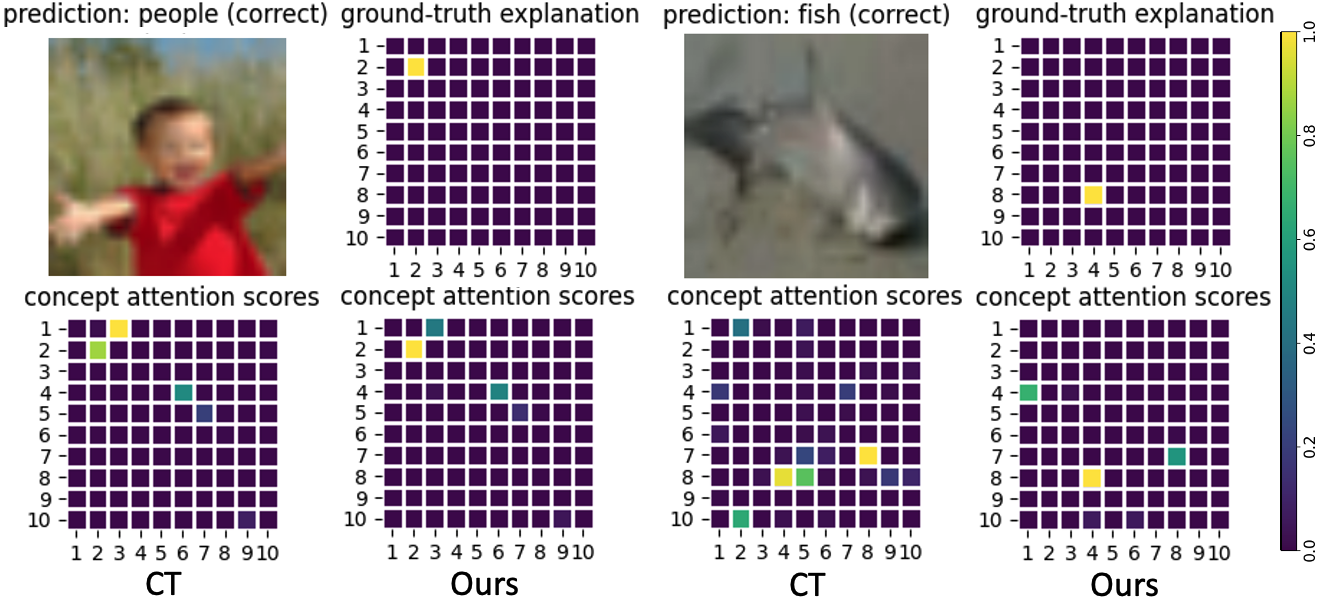}
    \caption{Comparison of class predictions for CT~\protect\cite{rigotti2021attention} and CCT (``Ours'') in examples where both make correct CIFAR100 super-class predictions. The 100 classes are indexed from 1~(top left in 10x10~grid) to 100~(bottom right in 10x10~grid). (\textbf{Left}) Ground-truth class label is \emph{boy}~(12), but CT mispredicted as \emph{baby}~(3), whereas CCT's prediction is correct. (\textbf{Right}) Ground-truth label is \emph{shark}~(84), but CT incorrectly selects \emph{ray}~(78) whereas CCT again makes a correct class prediction.}
    \label{fig:cifar100_exp_01}
\end{figure}

Figure~\ref{fig:cifar100_exp_01} shows two examples that demonstrate where our CCT's concept learning excels both quantitatively and qualitatively over CT.
CT's poor fine-grained class accuracy might be the result of making \emph{hallucinations} to greedily achieve super-class accuracy without forming sound latent concepts. Though both models correctly classify super-classes on the shown examples, the explanatory decision-making processes perform entirely differently in the two models. Although the ground-truth class of the left image in Fig.~\ref{fig:cifar100_exp_01} is \emph{boy}, the best-matching class concept with the highest attention score by CT was \emph{baby}, which is the incorrect fine-grained class but also belongs to the correct super-class \emph{people}. In the right example image, we observe a similar behavior of CT.
In contrast, our CCT's predicted concepts for both examples correctly match their ground-truth fine-grained classes with sparser concept attention scores than CT. Further details and results are described in Appendix~\ref{app_sub:cifar100}.

\subsection{Evaluation on CUB-200-2011}
\label{subsec:cub}

The CUB-200-2011~\cite{wah2011caltech} dataset comprises 11,788 bird images categorized into 200 species. Each image is annotated with various discrete concepts, e.g., the shape of the beak, or the color of the body, aiding species identification. The dataset involves 312 concepts distributed unevenly across images, so we utilize a pre-processing method from~\cite{rigotti2021attention} to address this. Additional results and details are described in Appendix~\ref{app_sub:cub}.

\paragraph{With Concept Explanations.}
We consider a real-world scenario where many-to-many and non-deterministic relationships between concepts and outputs exists, along with a mix of \emph{global} and \emph{spatially localized} concepts. We use CSA and CA modules within CCT to handle both global and spatial concepts, averaging their logits for interpretability. We use various backbones, including ViT~\cite{dosovitskiy2020image}, Swin Transformer~(SwinT)~\cite{liu2021swin}, and ConvNeXt~\cite{liu2022convnet} for this dataset. We use the embeddings of the tokenized image patches, while as input to the CCT in charge of the concepts, we use the embedding of the CLS token.

In Table~\ref{tab:eval_cub}, we compare our CCT with other methods based on \emph{Multi-stage}~(i.e., complex training) and \emph{End-to-end}~(i.e., training with backpropagation) training.
\begin{table*}[t!]
    \centering
    \begin{NiceTabular}{l|lll}
        Method & \multicolumn{3}{c}{Test Accuracy~($\%$)}\\
        \midrule\midrule
        \multirow{4}{1em}{Multi-stage} & Part R-CNN: $76.4$ & PS-CNN: $76.2$ & PN-CNN: $85.4$ \\    & SPDA-CNN: $85.1$ & PA-CNN: $82.8$ & MG-CNN: $83.0$ \\
        & 2-level attn.: $77.9$ & FCAN: $82.0$ & Neural const.: $81.0$ \\
        & ProtoPNet: $84.8$ &  Deform-ProtoPNet : $86.5$ & PIP-net : $84.3_{\pm 0.2}$\\
        \midrule
        \multirow{3}{1em}{End-to-end} & B-CNN: $85.1$ & CAM: $70.5$ & DeepLAC: $80.3$ \\
        & ST-CNN: $84.1$ & MA-CNN: $86.5$ & RA-CNN: $85.3$ \\
        & CEM: $77.1$ &  ProtoPFormer: $84.9$ & CT (w/w.o): $86.4_{\pm 0.2}/75.4_{\pm 0.3}$ \\
        \midrule
        \multirow{3}{1em}{CCT (ours)} & ViT-L+SA: $90.0_{\pm 0.3}$ & ViT-L+I-SA: $90.3_{\pm 0.3}$ & ViT-L+BO-QSA: $90.3_{\pm 0.1}$ \\
        & SwinT-L+SA: $90.7_{\pm 0.02}$ & SwinT-L+I-SA : $90.9_{\pm 0.4}$ & SwinT-L+BO-QSA (w/w.o): $\mathbf{91.2}_{\pm 0.2} / 90.9_{\pm 0.4}$ \\
        & ConvNeXt-L+SA: $87.8_{\pm 0.3}$ & ConvNeXt-L+I-SA: $89.3_{\pm 0.6}$ & ConvNeXt-L+BO-QSA : $89.4_{\pm 0.4}$ \\
    \bottomrule
    \end{NiceTabular}
    \caption{Performance comparison on CUB-200-2011. For B-CNN~\cite{lin2015bilinear}, Part R-CNN~\cite{zhang2014part}, PS-CNN~\cite{huang2016part}, PN-CNN~\cite{branson2014bird}, SPDA-CNN~\cite{zhang2016spda}, PA-CNN~\cite{krause2015fine}, MG-CNN~\cite{wang2015multiple}, 2-level attn.~\cite{xiao2015application}, FCAN~\cite{liu2016fully}, Neural const.~\cite{simon2015neural}, ProtoPNet~\cite{chen2019looks}, CAM~\cite{zhou2016learning}, DeepLAC~\cite{lin2015deep}, ST-CNN~\cite{jaderberg2015spatial}, MA-CNN~\cite{zheng2017learning}, and RA-CNN~\cite{fu2017look}, 
    performance from~\cite{rigotti2021attention}.
    For ProtoPFormer~\cite{xue2022protopformer} and CEM~\cite{zarlenga2022concept}, best performance directly from their works.
    For CT~\cite{rigotti2021attention} and CCT, results from our evaluation. The number of parameters of the backbone we use is: 307M~(ViT-L), 197M~(SwinT-L), and 197M~(ConvNeXt-L). (w/w.o) indicates the performance with/without ground-truth concept explanations.}
    \label{tab:eval_cub}
\end{table*}
All of the configurations of our CCT achieve over $87$\% classification accuracy, clearly outperforming other approaches. This confirms that the overall configuration of CCT enhances classification performance. Notably, our model surpasses non-interpretable baselines~(B-CNN) and methods requiring complex training.

\begin{figure}[t!]
    \centering
    \includegraphics[width=0.48\textwidth]{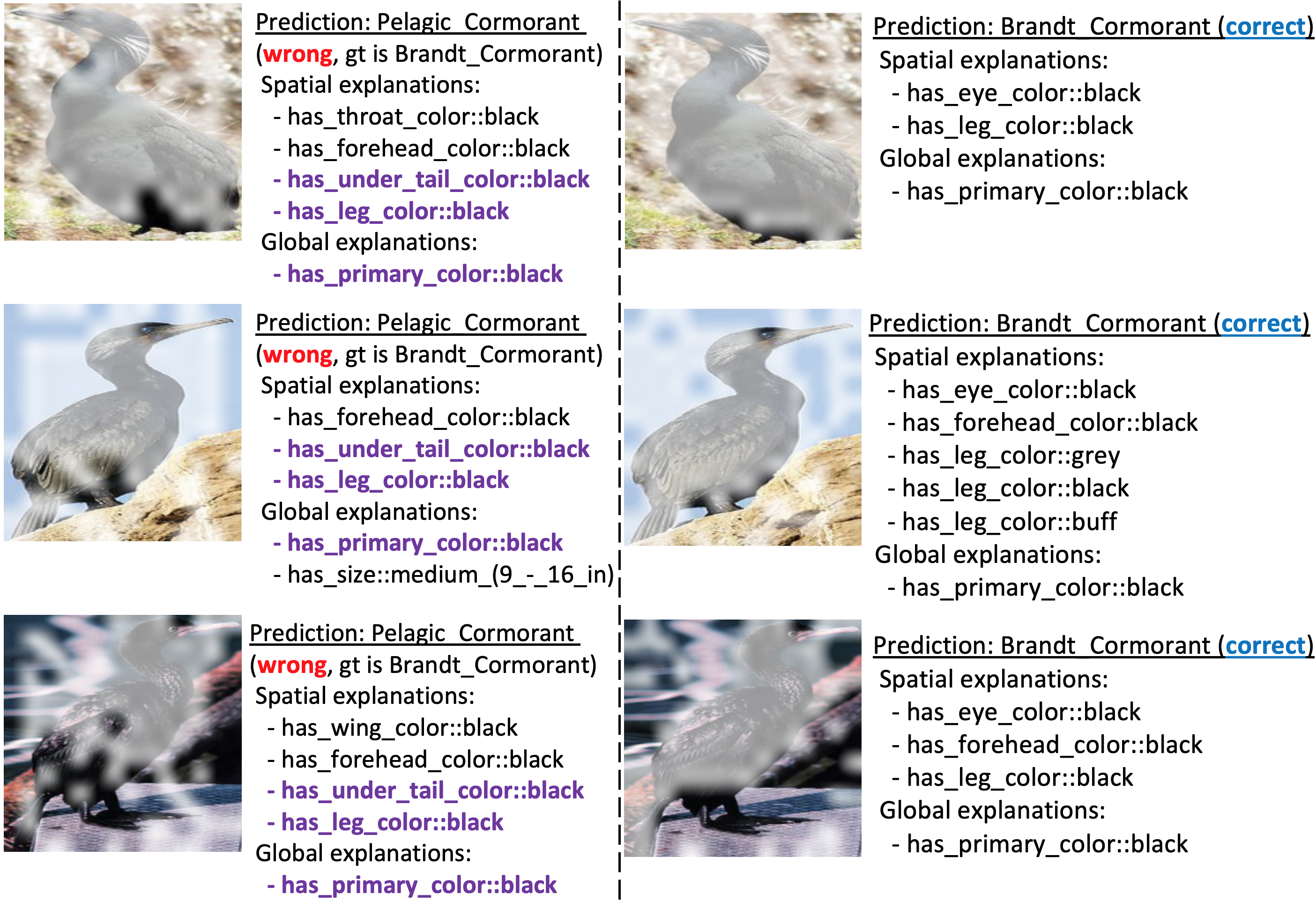}
    \caption{Prediction comparison between CT and our CCT with ground-truth explanation. (\textbf{Left}) All CT's predictions are incorrect. The highlighted explanations in purple are key attributes typically present when correctly classifying \texttt{Pelagic\_Cormorant}, the incorrect label. (\textbf{Right}) Our CCT's predictions are correct.}
    \label{fig:cub_exp_01}
\end{figure}

Figure~\ref{fig:cub_exp_01} highlights the distinction between CT and our CCT, illustrating that CT lacks learning global concepts. We emphasize concepts with the highest attention scores in all images. The figure shows a case where the CT's predictions were completely incorrect by converging to the single wrong label. Although the ground-truth class of the images was \texttt{Brandt\_Cormorant}, all CT's predictions were \texttt{Pelagic\_Cormorant}. Furthermore, the attributes CT used to make the incorrect classification included a subset of attributes~(purple-colored attributes in Fig.~\ref{fig:cub_exp_01}) typically associated with correct predictions of the incorrect \texttt{Pelagic\_Cormorant} label. Thus, the image-patch-centric CT hallucinates key aspects associated with incorrect labels, whereas our CCT shows more robust concept explanations of correct classifications.

\paragraph{Without Concept Explanations.}
Although the domain expert’s knowledge is the most effective tool for guiding a model’s explainability, the pre-precessing step to define the visual concepts for tasks may require time-consuming labeling and rely on human judgment. Importantly, we demonstrate that our CCT also works effectively without concept explanations, a capability not shared by other models. We can easily set up our loss with $\lambda_{expl} = 0$~(Final Loss defined in section~\ref{subsec:training}) and evaluate our model as the same hyperparameter setups of the experiment with explanation using only classification loss and sparsity loss.

In Table~\ref{tab:eval_cub}, our CCT's best configuration~(Swin-L+BO-QSA) without explanation achieved $90\%$ test accuracy, which is very marginal compared to the one with explanation. In contrast, CT can also be trained without explanation, but its performance is starkly degraded. 

\begin{figure}[t!]
    \centering
    \includegraphics[width=0.47\textwidth]{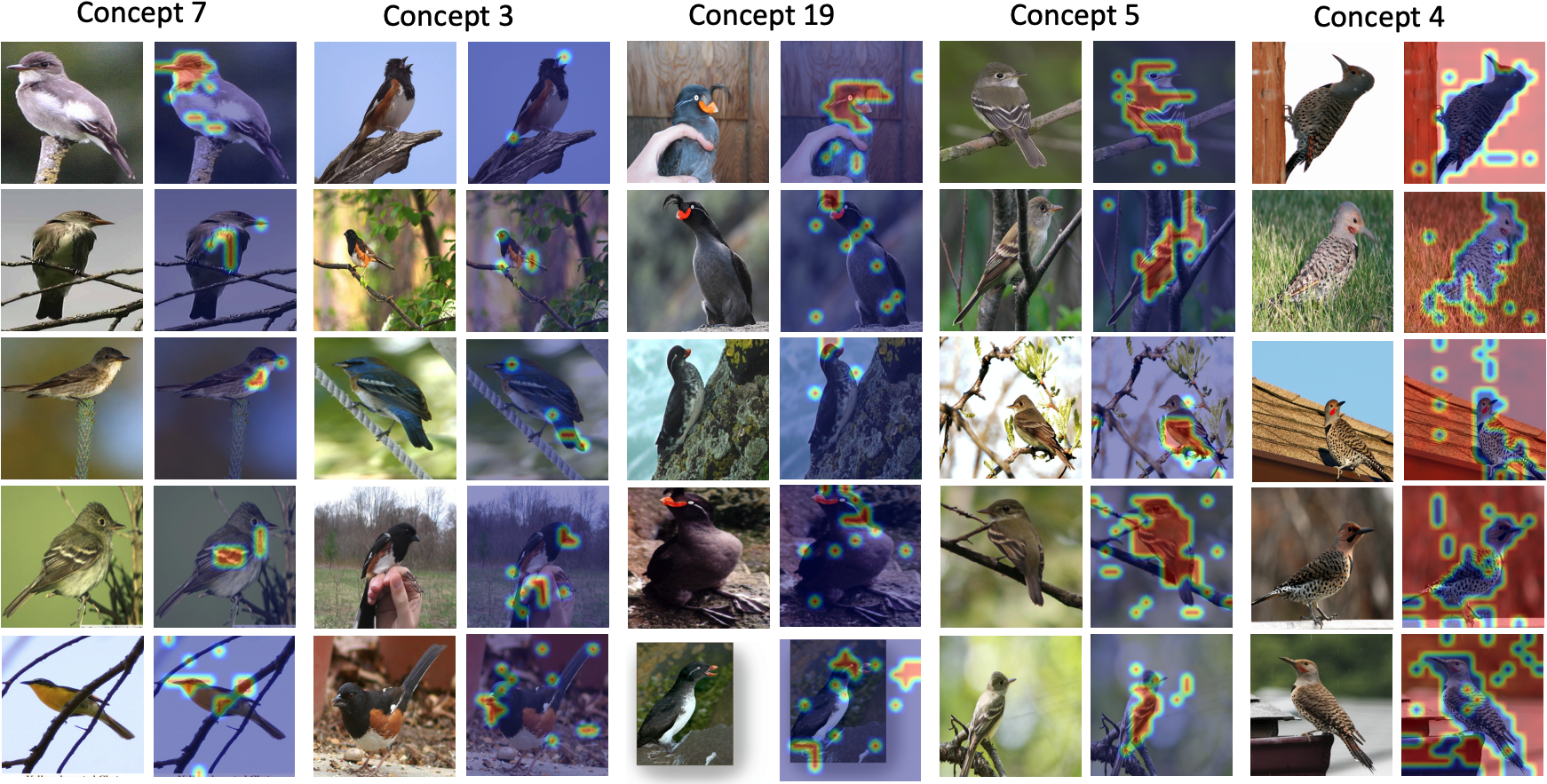}
    \caption{The activation of latent concepts learned from CUB-200-2011 and examples showing each latent concept. Further results can be found in Appendix~\ref{app_sub:cub}.}
    \label{fig:cub_concepts}
\end{figure}

Figure~\ref{fig:cub_concepts} visualizes the activation of the learned latent concepts in our model from the dataset.
Following~\cite{wang2023learning}, we additionally trained our CCT by setting the number of class labels~($n$) to 50 and the number of latent concepts~($k$) to 20. The figure showcases five latent concepts---7, 3, 19, 5, and 4. In our experimental results, we identified key concepts that CCT focuses on when classifying bird images. Concept 7 focuses on the beak and upper torso, while Concept 3 considers multiple features like the beak, eyes, and tail. Concept 19 captures unique head and feather structures, Concept 5 outlines the bird's entire body, and Concept 4 isolates birds from complex backgrounds. These demonstrate the model's nuanced focus and classification skills, even in an unsupervised environment where we don't have access to its concept explanation.


\subsection{Evaluations on ImageNet}
\label{subsec:imagenet}

Finally, to validate that our model can learn latent concepts without explanations, we tested CCT on ImageNet following the approach in~\cite{wang2023learning}. We used the relatively small ViT~(ViT-S) as the backbone\footnote{We avoided using a backbone larger than ResNet-101~(45M), which excluded SwinT-S~(50M) and ConvNeXt-S~(50M).}. Additional results and details are in Appendix~\ref{app_sub:imagenet}.

Table~\ref{tab:eval_imagenet} shows ImageNet classification performance.
\begin{table}[t!]
    \centering
    \begin{threeparttable}
        \begin{NiceTabular}{l|c}
            Model &  Test Acc.~($\%$) \\
            \midrule\midrule
            Vanilla ViT-S & $83.3_{\pm 0.2}$ \\
            \midrule
            ProtoPFormer(Deit-B)~\cite{xue2022protopformer} & $83.4_{\pm 2.2}$\\
            ProtoPool~\cite{DBLP:journals/corr/abs-2112-02902}~(ResNet-101) & $76.5_{\pm 0.8}$ \\
            ProtoPNet~\cite{DBLP:journals/corr/abs-1806-10574}~(ResNet-101) & $77.7_{\pm 0.3}$ \\
            Deform-ProtoPNet~\cite{DBLP:journals/corr/abs-2111-15000}~(ResNet-101)  & $76.1_{\pm 0.3}$ \\
            CT~\cite{rigotti2021attention}~(ViT-S) & $27.0_{\pm 0.2}$ \\
            $\text{BotCL}^{\dag}$~\cite{wang2023learning}~(ResNet-101) & $83.0$ \\
            \midrule
            CCT: ViT-S+SA & $76.3_{\pm 0.2}$ \\
            CCT: ViT-S+I-SA & $83.6_{\pm 0.2}$ \\
            CCT: ViT-S+BO-QSA & $\textbf{83.7}_{\pm 0.2}$ \\
        \bottomrule
        \end{NiceTabular}
    \end{threeparttable}
    \caption{Test accuracy on ImageNet. We used the first 200 classes following~\cite{wang2023learning}. $\dag$ indicates the best result from~\cite{wang2023learning}. The number of parameters is: 22M~(ViT-S), 45M~(ResNet-101), 50M~(ConvNeXt-S), and 86M~(Deit-B).}
    \label{tab:eval_imagenet}
\end{table}
The combination of ViT-S with BO-QSA in the experiment has the best performance, achieving $83.7\%$ test accuracy despite using a small-sized backbone, which meets or surpasses the performance of other SOTA methods---including some concept-based approaches that were not applicable or achieved very poor performance. 

\begin{figure}[t!]
    \centering
    \includegraphics[width=0.47\textwidth]{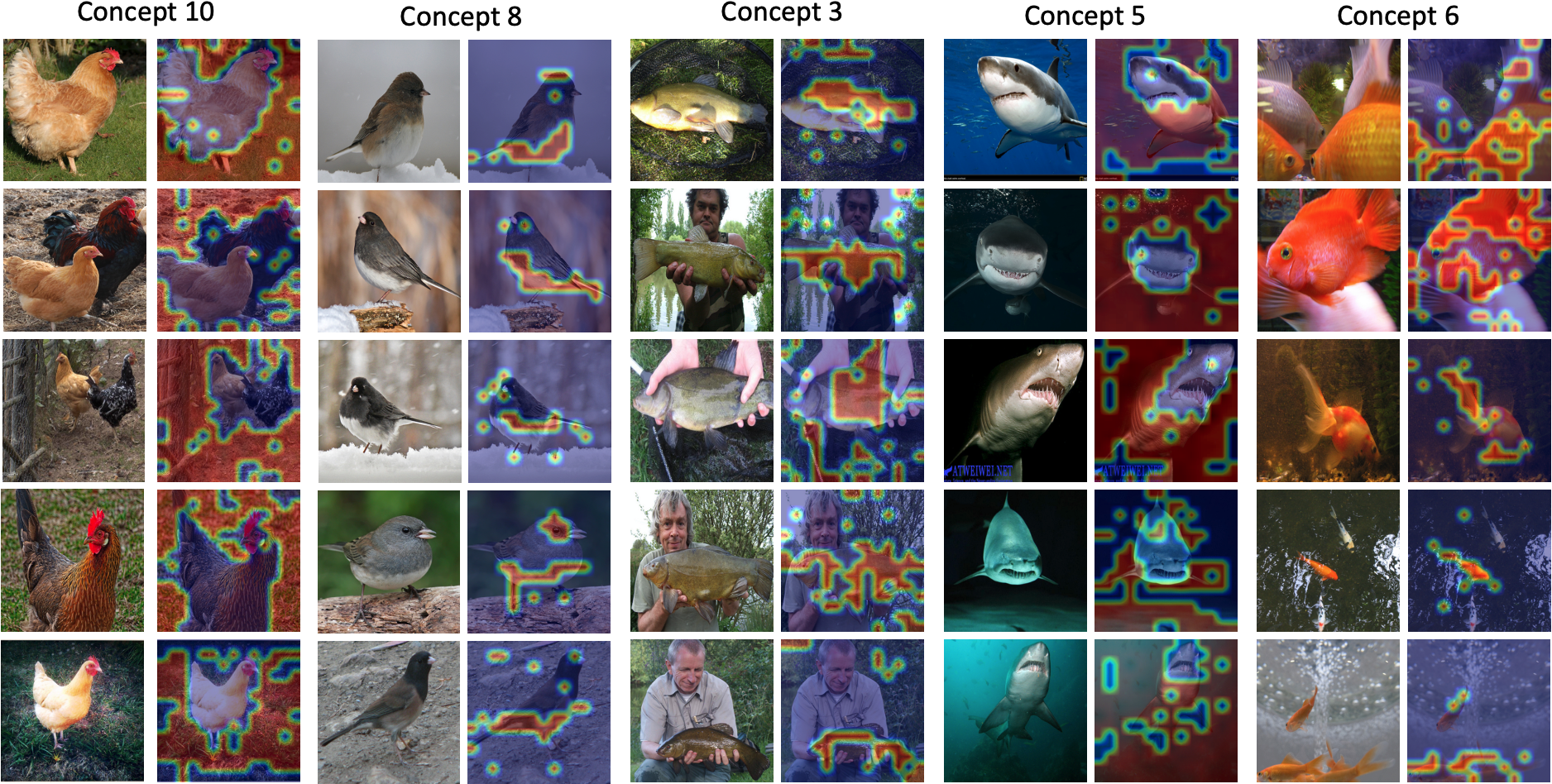}
    \caption{The activation of latent concepts learned from ImageNet and examples showing each latent concept. Further results can be found in Appendix~\ref{app_sub:imagenet}.}
    \label{fig:imagenet_concepts}
\end{figure}

In addition, we trained a simple CCT model by setting the number of classes~($n$) to 20 and the number of latent concepts~($k$) to 10 as in~\cite{wang2023learning} to visualize the consistency of the learned concepts. Figure~\ref{fig:imagenet_concepts} depicts five concepts we selected and five pairs of representative images for each latent concept learned from ImageNet.
Our CCT shows that it excels in isolating and emphasizing specific object features across various concepts. Concept 10 isolates hen contours by highlighting the background, while Concept 8 focuses on birds' ventral regions. Concept 3 skillfully segregates fish components, Concept 5 captures shark oral regions, and Concept 6 outlines goldfish shapes. CCT's capabilities in contouring and highlighting semantically meaningful areas surpass those of existing models. It also excels in unsupervised image retrieval, clustering semantically similar images together as evident in Fig.~\ref{fig:imagenet_concepts}. This showcases the strength of our concept-centric approach in generating semantically coherent results.

\section{Conclusions}
\label{sec:conclusions}
We proposed Concept-Centric Transformers, an intrinsically interpretable model via a Shared Global Workspace, allowing for achieving better interpretability, performance, and versatility for when expert knowledge is available or not.
A natural future research direction is to extend the concept extraction module to acquire composable ``pieces'' of knowledge~\cite{singh2023neural} and then learn the underlying composition rules or mechanisms relating the acquired pieces to each other. Those rules can be captured formally, as with first-order logic~\cite{barbiero2022entropy}. It is also promising to investigate other applications, such as model debugging and medical-image diagnosis.

\paragraph{\textbf{Acknowledgments:}} Supported in part by NSF award 2223839 and USACE ERDC award W912HZ-21-2-0040.

{\small
\bibliographystyle{ieee_fullname}
\bibliography{main}
}

\clearpage

\appendix
\section*{Supplementary Material}
\label{sec:appendix}
\renewcommand\thefigure{A-\arabic{figure}}\setcounter{figure}{0}
\renewcommand\thetable{A-\arabic{table}}
\setcounter{table}{0}
\setcounter{section}{0}
\pagenumbering{Alph}

\section{Reproducibility}
\label{app:reprod}
All source codes, figures, models, etc., are available at 
\url{https://github.com/PavlicLab/WACV2024-Hong-Concept_Centric_Transformers.git}.

\section{Method Details}
\label{app:method_detail}

\paragraph{Algorithm.}
Algorithm~\ref{alg:csa_module} shows the details of the Concept-Slot-Attention~(CSA) module in our Cocenpt-Centric Transformer~(CCT) in pseudo-code. Algorithm~\ref{alg:csa_module} is described based on the SA~\cite{locatello2020object}, but we simplify it by removing the last \texttt{LayerNorm} and \texttt{MLP} layers. Because of the modular characteristics in our framework, we leverage three slot-based approaches, including SA~\cite{locatello2020object}, I-SA~\cite{chang2022object}, and BO-QSA~\cite{jia2022improving}, which are interchangeable. We implemented the above approaches based on their official repositories/works.
\begin{algorithm}[b!]
\caption{\textbf{Concept-Slot-Attention~(CSA) module}. The module receives the set of input features $\mathbf{E} \in \mathbb{R}^{L \times D}$; the number of concepts $C$; and the dimension of concepts $d$. The model parameters include: the linear projection $q^{\mathtt{CSA}}, k^{\mathtt{CSA}}, v^{\mathtt{CSA}}$ with output dimension $d$; a \texttt{GRU} network; a Gaussian distribution's mean and diagonal covariance $\mu, \sigma \in \mathbb{R}^{d}$.}
\label{alg:csa_module}
\begin{algorithmic}
\State $\mathbf{S}^{\mathtt{concept}} = \mathtt{Tensor}(C, d)$
\Comment{$\mathtt{concept-slots} \in \mathbb{R}^{C \times d}$}
\State $\mathbf{S}^{\mathtt{concept}} \sim\mathcal{N}(\mu, \sigma)$
\State $\mathbf{E} = \mathtt{LayerNorm}(\mathbf{E})$
\For{$t = 0, \dots, T$}
    \State $\mathbf{S}^{\mathtt{concept}} = \mathtt{LayerNorm}(\mathbf{S}^{\mathtt{concept}})$
    \State $\mathbf{A}^{\mathtt{CSA}} = \operatorname{softmax}(\frac{1}{\sqrt{d}}q^{\mathtt{CSA}}(\mathbf{S}^{\mathtt{concept}}) \cdot k^{\mathtt{CSA}}(\mathbf{E})^{\top}, \mathtt{axis='concept-slots'})$
    \State $\mathbf{A}^{\mathtt{CSA}} = \mathbf{A}^{\mathtt{CSA}}\;/\;\mathbf{A}^{\mathtt{CSA}}\mathtt{.sum(axis='inputs')}$
    \State $\mathbf{U} = \mathbf{A}^{\mathtt{CSA}} \cdot v^{\mathtt{CSA}}(\mathbf{E})$
    \State $\mathbf{S}^{\mathtt{concept}} = \mathtt{GRU}(\mathtt{state} = \mathbf{S}^{\mathtt{concept}}, \mathtt{inputs} = \mathbf{U})$
\EndFor \\
\Return $\mathbf{S}^{\mathtt{concept}}$
\end{algorithmic}
\end{algorithm}

\paragraph{Positional Embedding for Concept Slots.} Because the resulting set of concept slots is orderless~\cite{locatello2020object}, it is difficult for the later module to: (i)~recognize which high-level concept each concept slot is representing and (ii)~identify which concept slot each high-level concept belongs to. Thus, we add a positional encoding $\mathbf{p}_{c}$ to each concept slot to avoid these challenges; in particular, $\Hat{\mathbf{s}}_{c} = \mathbf{s}_{c} + \mathbf{p}_{c}$.
After this, the concept embedding representation $\Hat{\mathbf{S}}^{\mathtt{concept}} \in \mathbb{R}^{C \times d}$ with positional embedding $\mathbf{P}$ is passed as the final concept representation to the next module.

\paragraph{Number of Iterations in the Concept Slot Attention module.}
In the original usage of slot-based approaches, the refinement could be repeated several times depending on the tasks.
For setting the number of iterations $T$~(Algorithm~\ref{alg:csa_module}), we set the different number of iterations $T$ for the three slot-based methods we used. For the vanilla SA, we set $T$ to 1 because the initialization for slots in the SA has some limitations revealed by~\cite{chang2022object}, and we found that a single iteration to update the concept slots benefits in which the learned concept slots possess much information is performed through the training with backpropagation. In other words, our proposed interpretable broadcast scheme~(\ref{subsec:broadcast} in the main text) contributes to better performances for the SA than the internally refined iterations used in the conventional slot-based methods. Furthermore, we set $T$ to $3$ for the BO-QSA and the I-SA by simply following the best values of iterations in their works.  

\paragraph{Comparison with Concept Transformers~\cite{rigotti2021attention}.}

As shown in~\cite[Fig.~1]{rigotti2021attention}, Concept Transformers~(CTs) are the limited usage in our formulation, which leverages learnable vectors for concept learning instead of using ``a shared workspace''. We highlight that the concept embedding representation in our CCT is more sophisticated and generalizable than the one in CTs. Each concept embedding in CT is represented as a simple learnable vector shared with all input batches to learn. As such, it may be difficult for the vector to capture which image features can contribute to each concept more than others. For instance, as shown in Fig.~\ref{fig:overall_architecture} in the main text, given three images belonging to the same class ``Winter Wren'', the image features with the same spatial positions from each image can have different importance for providing information to learn the same global concept, such as \emph{``what is the main body color for Winter Wren?''} or \emph{``what is the body shape for it?''}. In contrast, our proposed module naturally aggregates how much each image feature contributes to each concept using the attention $\mathbf{A}^{\mathtt{CSA}}$~(See in Algorithm~\ref{alg:csa_module}) and provides batch-specific concept embedding representations that have more semantically meaningful information. In all of the experiments, we demonstrate that our CCTs always outperform CTs.

\section{Further Experimental Results and Details}
\label{app:exp}
In this section, we explain further experimental results and details. All experiments are conducted with three different random seeds and $95\%$ confidence intervals. 

\subsection{Dataset Statistics}
Table~\ref{tab:dataset_stat} depicts the statistics of all benchmark datasets in our experiments. Because all datasets have no portion of validation, we manually pick the portion of the validation dataset for exploring the best hyperparameters for models. 

For CIFAR-100 Super-class dataset, it serves as a notable example for nuanced image categorization. Originating from the original CIFAR-100 dataset, it contains 100 fine-grained image classes that are aggregated into 20 distinct super-classes for broader categorization. For instance, the super-class "vehicles 1" comprises specific, fine-grained classes like "bicycle," "bus," "motorcycle," "pickup truck," and "train." This structured hierarchy makes the dataset a widely used benchmark in evaluating deep learning models, particularly those that incorporate logical constraints. 

For CUB-200-2011, we explain the pre-processing steps following~\cite{rigotti2021attention}. Initially, the dataset has 312 binary attributes, but we filter to retain only those attributes that occur in at least 45$\%$ of all samples in a given class and occur in at least 8 classes. Thus, we get a total of 108 attributes, and based on this, we group them into two kinds of concepts: spatial and global concepts. We finally get 13 global and 95 spatial concepts by looking at each attribute. For example, \texttt{has\_shape::perching-like} and \texttt{has\_primary\_color::black} are global concepts, and \texttt{has\_eye\_color::black} and \texttt{has\_forehead\_color::yellow} are spatial concepts.

For the ImageNet dataset, we adopted the methodology presented in~\cite{wang2023learning} to validate our CCT's ability to learn latent concepts without the need for explicit concept explanations. \cite{wang2023learning} uses attention maps to identify a predefined set of concepts within the dataset, focusing on the first 200 classes of ImageNet for their evaluation. Following this approach, we also utilized the first 200 classes in ImageNet for training and conducted evaluations to determine test accuracy on the ImageNet dataset. This experiment serves to benchmark our CCT model's capabilities in learning latent concepts without explicit explanations, further showcasing its effectiveness in an unsupervised setting.
\begin{table*}[t!]
\caption{Benchmark Dataset Statistics. $\dagger$ indicates the rescaled size of inputs for the ViT backbone, which is different from the original sizes of the datasets. For the ImageNet experiment setup, we follow~\cite{wang2023learning}.}
\centering
\begin{NiceTabular}{m{0.22\textwidth}|m{0.21\textwidth}|m{0.24\textwidth}|m{0.24\textwidth}}
\textbf{Dataset} & \textbf{CIFAR100 Super-class} & \textbf{CUB-200-2011} & \textbf{ImageNet} \\
\midrule\midrule
Input size & $3 \times 28 \times 28$ & $3 \times 224 \times 224^{\dagger}$ & $3 \times 224 \times 224^{\dagger}$ \\ 
\midrule
\# Classes & 20~(super-class) & 200~(bird species) & 200~(objects) \\ 
\midrule
\# Concepts & 100~(class) & 13~(global), 95~(spatial) & 50~(latent spatial) \\
\midrule
\# Training samples & 55,000 & 5,994 & 255,224 \\ 
\midrule
\# Validation samples & 5,000 & 1,000 & 10,000 \\
\midrule
\# Test samples & 10,000 & 4,794 & 20,000 \\
\bottomrule
\end{NiceTabular}
\label{tab:dataset_stat}
\end{table*}

\subsection{Hardware Specification of The Server}
The hardware specification of the server that we used to experiment is as follows:
\begin{itemize}
    \item CPU: Intel\textregistered{} Core\textsuperscript{TM} i7-6950X CPU @ 3.00GHz (up to 3.50 GHz)
    \item RAM: 128 GB (DDR4 2400MHz)
    \item GPU: NVIDIA GeForce Titan Xp GP102 (Pascal architecture, 3840 CUDA Cores @ 1.6 GHz, 384-bit bus width, 12 GB GDDR G5X memory)
\end{itemize}

\subsection{Model Architectures}
We leverage three kinds of backbones, including Vision Transformers~(ViT)~\cite{dosovitskiy2020image}, Swin Transformers~(Swin)~\cite{liu2021swin}, and ConvNeXt~\cite{liu2022convnet}. We employ \href{https://huggingface.co/docs/timm/index}{\texttt{timm}} Python library supported by Hugging Face\textsuperscript{\texttrademark}.

\paragraph{Variants of ViT.}
In our experiments, we use three variants of Vision Transformer~(ViT), \texttt{ViT-Large}, \texttt{ViT-Small}, and \texttt{ViT-Tiny}~(\texttt{vit\_large\_patch16\_224}, \texttt{vit\_small\_patch16\_224}, and \texttt{vit\_tiny\_patch16\\\_224}, respectively in \href{https://huggingface.co/docs/timm/index}{\texttt{timm}}). These variants are defined by their number of encoder blocks, the number of attention heads on each block, and the dimension of the hidden layer. The \texttt{ViT-Large} has 24 encoder blocks with 16 heads, and the dimension of the hidden layer is 1024. In addition, The \texttt{Vit-Small} has 12 encoder blocks with 6 heads, and the dimension of the hidden layer is 384. Finally, the \texttt{ViT-Tiny} has 12 encoder blocks with 3 heads, and the dimension of the hidden layer is 192, which is much more lightweight. A comparison between the variants of the ViT is shown in Table~\ref{tab:vit_variants}. 

\paragraph{Architecture of Swin-Large.}
In our experiment, we use the Swin Transformer~(Swin)-Large~(\texttt{swin\_large\_patch4\_window7\_224\\.ms\_in22k} in \href{https://huggingface.co/docs/timm/index}{\texttt{timm}}). See Table~\ref{tab:Swin_variants} for the overall architecture of Swin-Large.

\paragraph{Architecture of ConvNeXt-Large.}
In our experiment, we leverage ConvNeXt-Large~(\texttt{convnext\_large\\.fb\_in22k\_ft\_in1k} in \href{https://huggingface.co/docs/timm/index}{\texttt{timm}}). See Table ~\ref{tab:ConvNeXt_variants} for the overall architecture of ConvNeXt-Large. 
\begin{table*}[t!]
    \centering
    \begin{NiceTabular}{c|c|c|c|c}
    Model & Layers & Hidden Dim. & Heads & \# Params. \\
    \midrule\midrule
    ViT-Tiny & 12 & 192 & 3 & 5M \\
    \midrule
    ViT-Small & 12 & 384 & 6 & 22M \\
    \midrule
    ViT-Large & 24 & 1024 & 16 & 304 M \\
    \bottomrule
    \end{NiceTabular}
    \caption{Comparison between variants of the ViT~\cite{dosovitskiy2020image}}
    \label{tab:vit_variants}
\end{table*}
\begin{table*}[t!]
    \centering
    \begin{NiceTabular}{c|c|c|c|c}
    Model & Layers ~(per stage)	 & Hidden Dim. & Attention Heads & \# Params. \\
    \midrule\midrule
    Swin-Large & ~(2, 2, 18, 2) & 192 & 6 & 197M \\
    \bottomrule
    \end{NiceTabular}
    \caption{Overall architecture of the Swin Transformer~\cite{liu2021swin}}
    \label{tab:Swin_variants}
\end{table*}
\begin{table*}[t!]
    \centering
    \begin{NiceTabular}{c|c|c|c|c}
    Model & Conv. Layers ~(per block)	 & Hidden Dim. & Attention Heads & \# Params. \\
    \midrule\midrule
    ConvNeXt-Large & 2 & 512 & 16 & 50M \\
    \bottomrule
    \end{NiceTabular}
    \caption{Overall architecture of the ConvNeXt~\cite{liu2022convnet}}
    \label{tab:ConvNeXt_variants}
\end{table*}

\subsection{CIFAR100 Super-class Experiments}
\label{app_sub:cifar100}
\paragraph{The Difference Between CCT and Other Deep-Learning Approaches with Logical Constraints.}
The logical constraint by~\cite{fischer2019dl2} to address this task was introduced, i.e., once the model classifies an image into a class, the predicted probabilities of that super-class must first arrive at the whole mass. Referring to this, several deep-learning approaches leverage it~\cite{hoernle2022multiplexnet, li2023learning}. For instance, the five classes, \emph{baby, boy, girl, man,} and \emph{woman}, belong to the super-class \emph{people}. Thus, $\text{Pr}_{\emph{people}}(\mathbf{x}) = 1$ if $\mathbf{x}$ is classified as \emph{baby}. So, logical constraints can be formulated as $\wedge_{s \in superclass} (\text{Pr}_{s}(\mathbf{x}) = 0 \vee \text{Pr}_{s}(\mathbf{x}) = 1)$, where the probability of the super-class is the sum of its corresponding classes' probabilities, e.g., $\text{Pr}_{\emph{people}}(\mathbf{x}) = \text{Pr}_{\emph{baby}}(\mathbf{x}) + \text{Pr}_{\emph{boy}}(\mathbf{x}) + \text{Pr}_{\emph{girl}}(\mathbf{x}) + \text{Pr}_{\emph{man}}(\mathbf{x}) +
\text{Pr}_{\emph{woman}}(\mathbf{x})$.

However, because our CCT~(and CT) follows Proposition 1 in~\cite{rigotti2021attention}, the probability of choosing the preferred superclass is $s^{c} = \arg\max_{s}(\beta_{c})_{s}$ of class $c$~(Refer to Eq.(\ref{eq:ca_prob}) in the main text). This means the super-class is determined by the largest attention score of the class among all classes, which is different from the definition of the logical constraint above. 

\begin{figure*}[t!]
    \centering
    \includegraphics[width=0.7\textwidth]{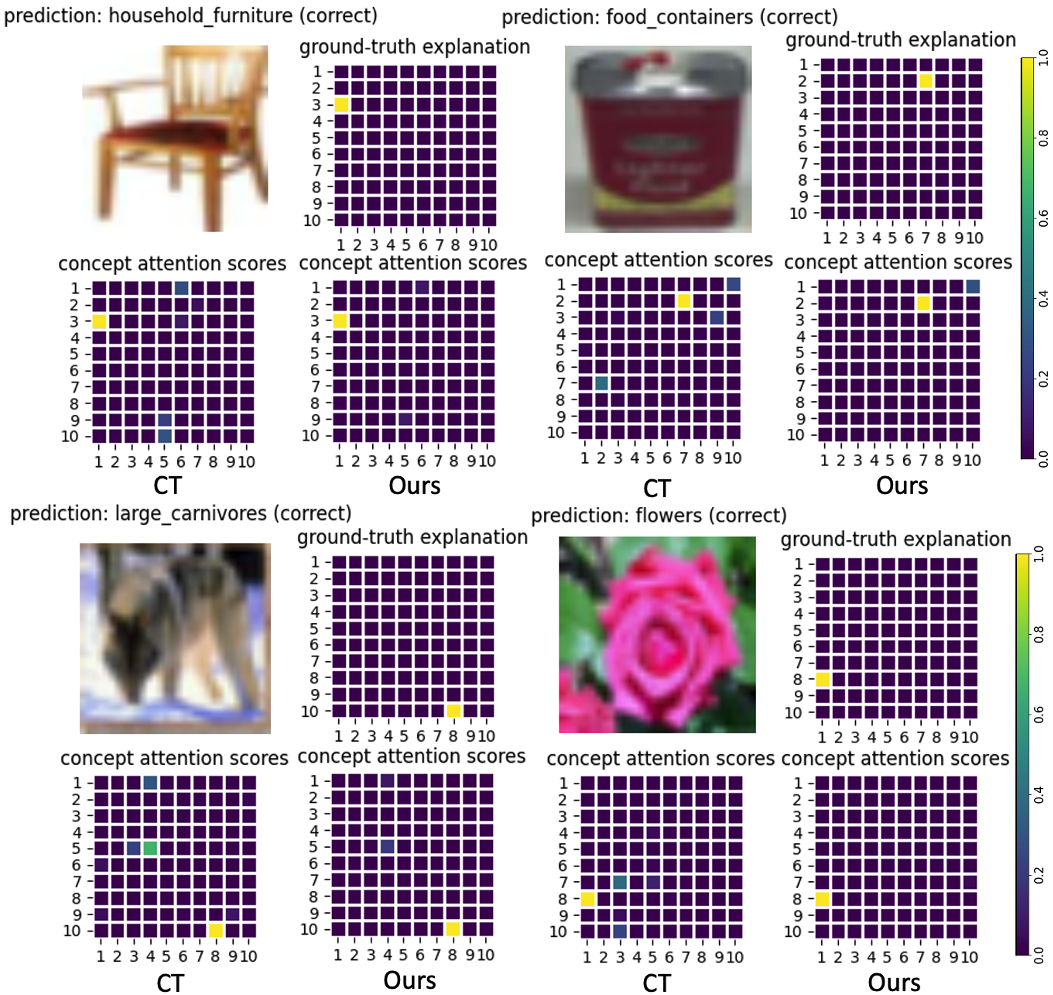}
    \caption{Examples of correct predictions by CT and CCT on CIFAR100 super-class. The 100 classes are indexed from 1~(top left in 10x10~grid) to 100~(bottom right in 10x10~grid). }
    \label{fig:cifar100_exp_02}
\end{figure*}

\paragraph{Additional Results.}
Figure~\ref{fig:cifar100_exp_02} showcases several examples of correct predictions by CT and our CCT, highlighting that compared to CT, our CCT achieves sparser explanation attention scores, which indicates a more confident and robust degree of belief for decision-making. 

%

\paragraph{Hyperparameter Settings.} 
For the experimental results, including Table~\ref{tab:eval_cifar100} and Fig.~\ref{fig:cifar100_exp_01} in the main text, referring to~\cite{fischer2019dl2}, we first find the best experimental setups of both ViT-Tiny and CT because they have not been evaluated on CIFAR100 Super-class before. Once the hyperparameter setup for the ViT-Tiny backbone is found to achieve similar performance to the Wide ResNet~(the backbone for the second baseline group), we apply the same hyperparameter setting to both CT and our CCT. Table~\ref{tab:hparam_cct_cifar100} presents all shared hyperparameters for ViT-Tiny, CT, and all configurations of CCT. Please refer to Table \ref{tab:hparam_Proto_archi} for hyperparameter settings used in our baseline experiments. 

\subsection{CUB-200-2011 Experiments}
\label{app_sub:cub}



\paragraph{Additional Results with Comparison with CT.}
With concept explanations, we introduce additional experimental results to compare predictions between CT and our CCT. 

\begin{figure*}[t!]
    \centering
    \includegraphics[width=0.9\textwidth]{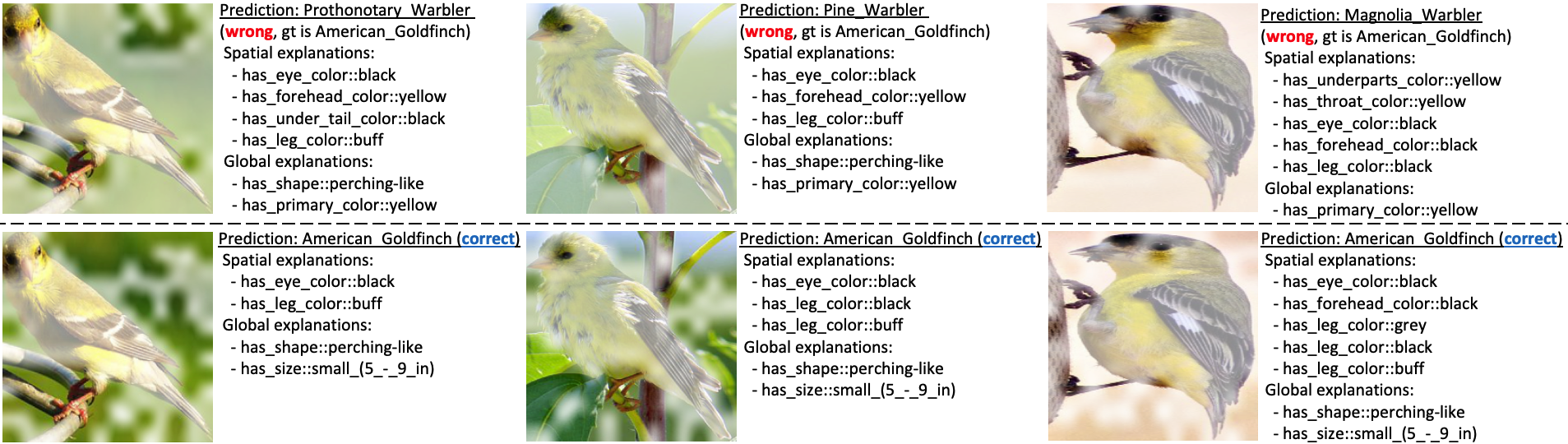}
    \caption{%
    Prediction comparison between CT and our CCT. (\textbf{Top}) CT's predictions are incorrect. (\textbf{Bottom}) All predictions are correct by our CCT.}
    \label{fig:cub_exp_02}
\end{figure*}

Figure~\ref{fig:cub_exp_02} depicts a case where the CT's predictions were utterly different from each other despite all images having the same class. The predicted classes from the CT in Fig.~\ref{fig:cub_exp_02} were \texttt{Prothonotary\_Warbler}, \texttt{Pine\_Warbler}, and \texttt{Magnolia\_Warbler}, respectively. The only difference between the first two predictions that the CT made is whether the spatial concept \texttt{has\_under\_tail\_color::black} exists, indicating the CT's performance sensitivity. In addition, the third prediction from the CT contains new spatial concepts unrelated to the first two examples, which shows analytical uncertainty as to why the CT made such a decision. In contrast, the predictions made by our CCT were all correct, with consistent global explanations. This indicates that although incorporating spatial concepts provide additional information, determining robust global concepts is more important for the classification task. 

\begin{figure*}[t!]
    \centering
    \includegraphics[width=0.9\textwidth]{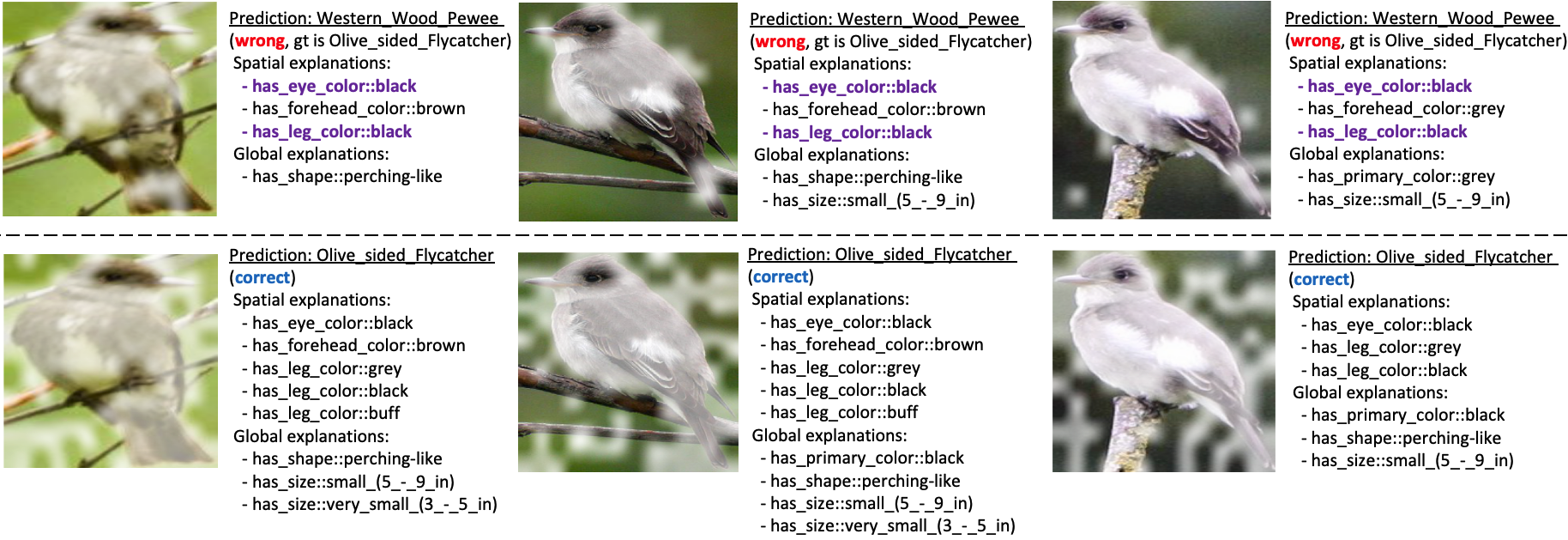}
    \caption{Prediction comparison between CT and CCT. (\textbf{Top Row}) CT's predictions are incorrect. The purple highlighted explanations are key attributes in \texttt{Western\_Wood\_Pewee}. (\textbf{Bottom Row}) All predictions are correct by our CCT.}
    \label{fig:cub_exp_03}
\end{figure*}

Figure~\ref{fig:cub_exp_03} shows some examples where our CCT outperforms CT to classify \texttt{Olive\_sided\_Flycatcher}. All CT's predictions were \texttt{Western\_Wood\_Pewee}, and this is caused by their spatial explanations, including \texttt{has\_eye\_color::black} and \texttt{has\_leg\_color::black}. We discovered that these two spatial attributes are critical for \texttt{Western\_Wood\_Pewee}, and thus mislead the model's predictions. In contrast, the explanations achieved by CCT are more robust and consistent. 

\begin{figure*}[t!]
    \centering
    \includegraphics[width=0.9\textwidth]{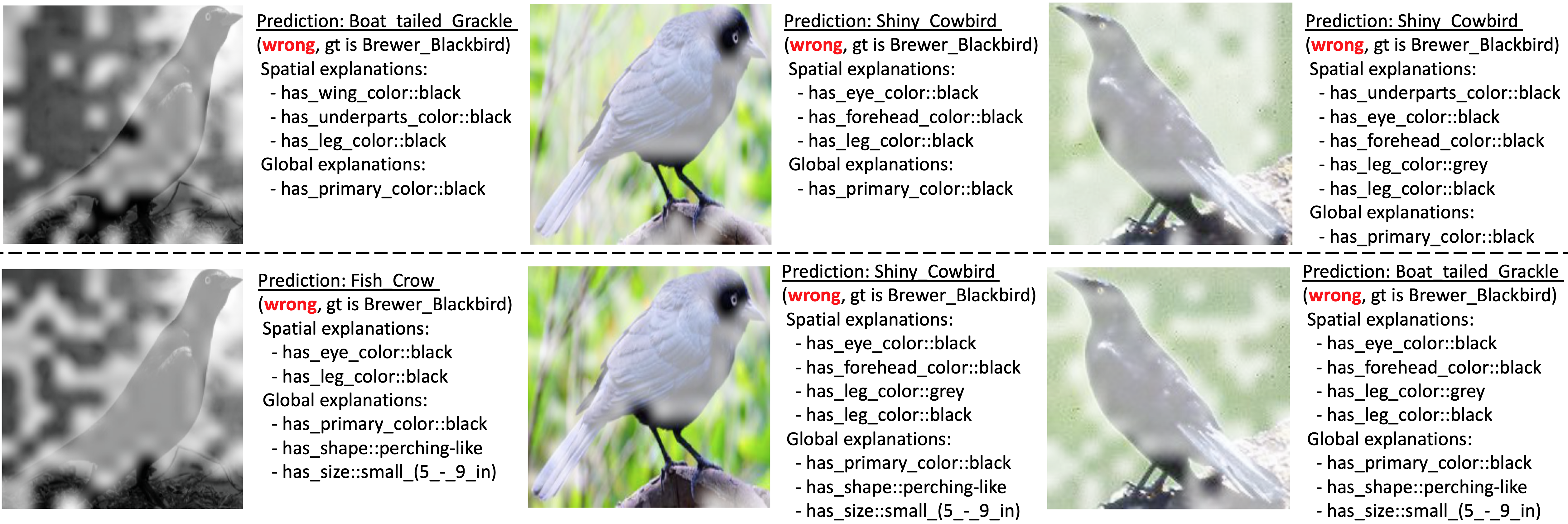}
    \caption{The example of failure predictions of both CT~(\textbf{Top Row}) and CCT~(\textbf{Bottom Row}).}
    \label{fig:cub_exp_fail_01}
\end{figure*}

\begin{figure*}[t!]
    \centering
    \includegraphics[width=0.9\textwidth]{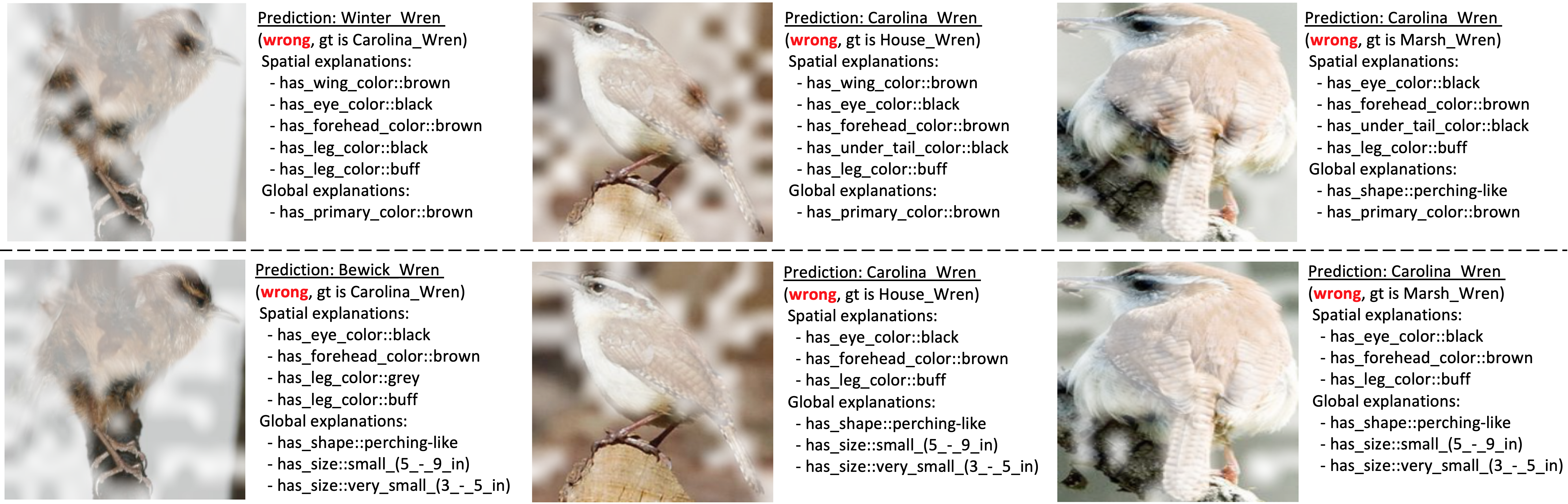}
    \caption{The example of failure predictions of both CT~(\textbf{Top Row}) and CCT~(\textbf{Bottom Row}).}
    \label{fig:cub_exp_fail_02}
\end{figure*}

Finally, Figs.~\ref{fig:cub_exp_fail_01} and~\ref{fig:cub_exp_fail_02} demonstrate the failure cases where both CCT and CT predict incorrectly. In both models, noise caused by spatial explanations common in some classes tends to impair the performance of prediction results. However, compared to CT, our proposed CCT captures richer global explanations, which thus helps to achieve better classification performance in general.

%
%

\paragraph{Hyperparameter Settings.} 
We follow the official implementation of CT and produce the experimental results, including Table~\ref{tab:eval_cub}, Fig.~\ref{fig:cub_exp_01}, in the main text, Fig.~\ref{fig:cub_exp_02}, and Fig.~\ref{fig:cub_exp_03} in Appendix. The only difference in hyperparameter settings between our CCT and CT is the learning rate for the AdamW optimizer, 5e-5 for CT, 1e-5 for the configurations with the ViT-Large backbone of our CCT, 2e-5 for our CCT with SwinT-Large, and 5e-5 for our CCT with ConvNeXt-Large. Table~\ref{tab:hparam_cub} shows the shared hyperparameters for both approaches. For those baselines not listed in Table~\ref{tab:hparam_cub}, we refer directly to the data presented in their respective publications.

\begin{table}[t!]
    \begin{subtable}[t!]{0.32\textwidth}
        \centering
        \begin{NiceTabular}{c|c}
        Name & Value \\
        \midrule
        Batch size & $64$ \\
        Epochs & $20$ \\
        Warmup Iters. & $10$ \\
        Learning rate & 5e-5 \\
        Explanation loss $\lambda$ & $1.0$ \\
        Weight decay & 1e-3 \\ 
        Attention sparsity & 0.5 \\
        \bottomrule
        \end{NiceTabular}
        \caption{ViT-Tiny, CT and all configurations of CCT on CIFAR100 Super-class}
        \label{tab:hparam_cct_cifar100}
     \end{subtable}
     \begin{subtable}[t!]{0.32\textwidth}
        \centering
        \begin{NiceTabular}{c|c}
        Name & Value \\
        \midrule
        Epochs & $60$ \\
        Learning Rate & 1e-4 \\
        Number of classes & $20$ \\
        Number of concepts & $10$ \\ 
        Quantity Bias & 0.1 \\
        Distinctiveness Bias & 0.05 \\
        Consistence Bias & 0.01 \\
        Weak Supervision Bias & 0.1 \\
        \bottomrule
        \end{NiceTabular}
        \caption{For botCL on CIFAR100 Super-class}
        \label{tab:hparam_botCL_cifar100}
     \end{subtable}
     \begin{subtable}[t!]{0.32\textwidth}
        \centering
        \begin{NiceTabular}{c|c}
        Name & Value \\
        \midrule
        Batch size & $16$ \\
        Epochs & $50$ \\
        Warmup Iters. & $10$ \\
        Explanation loss $\lambda$ & $1.0$ \\
        Weight decay & 1e-3 \\ 
        Attention sparsity & 0.5 \\
        \bottomrule
        \end{NiceTabular}
        \caption{For both CT and all configurations of CCT on CUB-200-2011}
        \label{tab:hparam_cub}
     \end{subtable}
     \begin{subtable}[t!]{0.32\textwidth}
        \centering
        \begin{NiceTabular}{c|c}
        Name & Value \\
        \midrule
        Batch size & $256$ \\
        Epochs & $10$ \\
        Warmup Iters. & $10$ \\
        Explanation loss $\lambda$ & $0.$ \\
        Weight decay & 1e-3 \\ 
        Attention sparsity & 0. \\
        \bottomrule
        \end{NiceTabular}
        \caption{For both CT and all configurations of CCT on ImageNet}
        \label{tab:hparam_cct_imagenet}
     \end{subtable}
     \caption{Shared hyperparameters on different datasets.}
     \label{tab:hparam}
\end{table}
\begin{table}[t!]
    \begin{subtable}[t!]{0.32\textwidth}  
        \centering
        \begin{NiceTabular}{c|c|c}
        Name & CIFAR100 & ImageNet \\
        \midrule
        Batch size & $64$ & $16$ \\
        Epochs & $10$ & $10$ \\
        Freeze Epochs & $10$ & $10$ \\
        Pre-train Epochs & $0$ & 0 \\
        Weight decay & $0.0$ & $0.0$ \\
        LR~(prototypes weights) & 5e-2 & 5e-2 \\
        LR~(backbone) & 5e-4 & 5e-4 \\
        \bottomrule
        \end{NiceTabular}
        \caption{For PIP-Net on CIFAR100 and ImageNet, LR stands for Learning Rate}
        \label{tab:hparam_imagenet_pipnet}
    \end{subtable}

    \begin{subtable}[t!]{0.32\textwidth}
        \centering
        \begin{NiceTabular}{c|c|c}
        Name & CIFAR100 & ImageNet \\
        \midrule
        Batch size & $64$ & $64$\\
        Epochs & $50$ & $10$\\
        Warmup Learning Rate & 1e-4 & 1e-4 \\
        Feature Learning Rate & 4e-4 & 4e-4 \\
        Prototype Learning Rate & 3e-3 &  3e-3\\ 
        Number of Prototype & $2000$ & $2000$ \\
        \bottomrule
        \end{NiceTabular}
        \caption{For ProtoPFormer on CIFAR100 and ImageNet}
        \label{tab:hparam_imagenet_ProtoPFormer}
     \end{subtable}

    \begin{subtable}[t!]{0.32\textwidth}
        \centering
        \begin{NiceTabular}{c|c|c}
        Name & CIFAR100 & ImageNet \\
        \midrule
        Batch size & $80$ & $80$\\
        Epochs & $30$ & $20$\\
        Learning Rate & 1e-3 & 1e-3 \\
        Gumbel Time & $30$ & $30$ \\
        Number of Classes & $20$ & $200$\\
        Number of Prototype & $202$ & $202$ \\
        Prototype Depth & $256$ & $256$ \\
        \bottomrule
        \end{NiceTabular}
        \caption{For ProtoPool on CIFAR100 and ImageNet}
        \label{tab:hparam_imagenet_ProtoPool}
     \end{subtable}

    \begin{subtable}[t!]{0.32\textwidth}
        \centering
        \begin{NiceTabular}{c|c|c}
        Name & CIFAR100 & ImageNet \\
        \midrule
        Batch size & $80$ & $80$\\
        Epochs & $100$ & $100$\\
        Learning Rates & default & default \\
        Number of Classes & $20$ & $200$\\
        Number of Prototype & $100$ & $200$ \\
        \bottomrule
        \end{NiceTabular}
        \caption{For Deformable-ProtoPNet on CIFAR100 and ImageNet}
        \label{tab:hparam_imagenet_Deform_ProtoPool}
     \end{subtable}

    \begin{subtable}[t!]{0.32\textwidth}
        \centering
        \begin{NiceTabular}{c|c|c}
        Name & CIFAR100 & ImageNet \\
        \midrule
        Batch size & $80$ & $80$\\
        Epochs & $100$ & $100$\\
        Learning Rates & default & default \\
        Number of Classes & $20$ & $200$\\
        Number of Prototype & $1000$ & $2000$ \\
        \bottomrule
        \end{NiceTabular}
        \caption{For ProtoPNet on CIFAR100 and ImageNet}
        \label{tab:hparam_imagenet_ProtoPNet}
     \end{subtable}
     \caption{Shared hyperparameters on different datasets.}
     \label{tab:hparam_Proto_archi}
\end{table}

\begin{figure*}[t!]
    \centering
    \includegraphics[width=\textwidth]{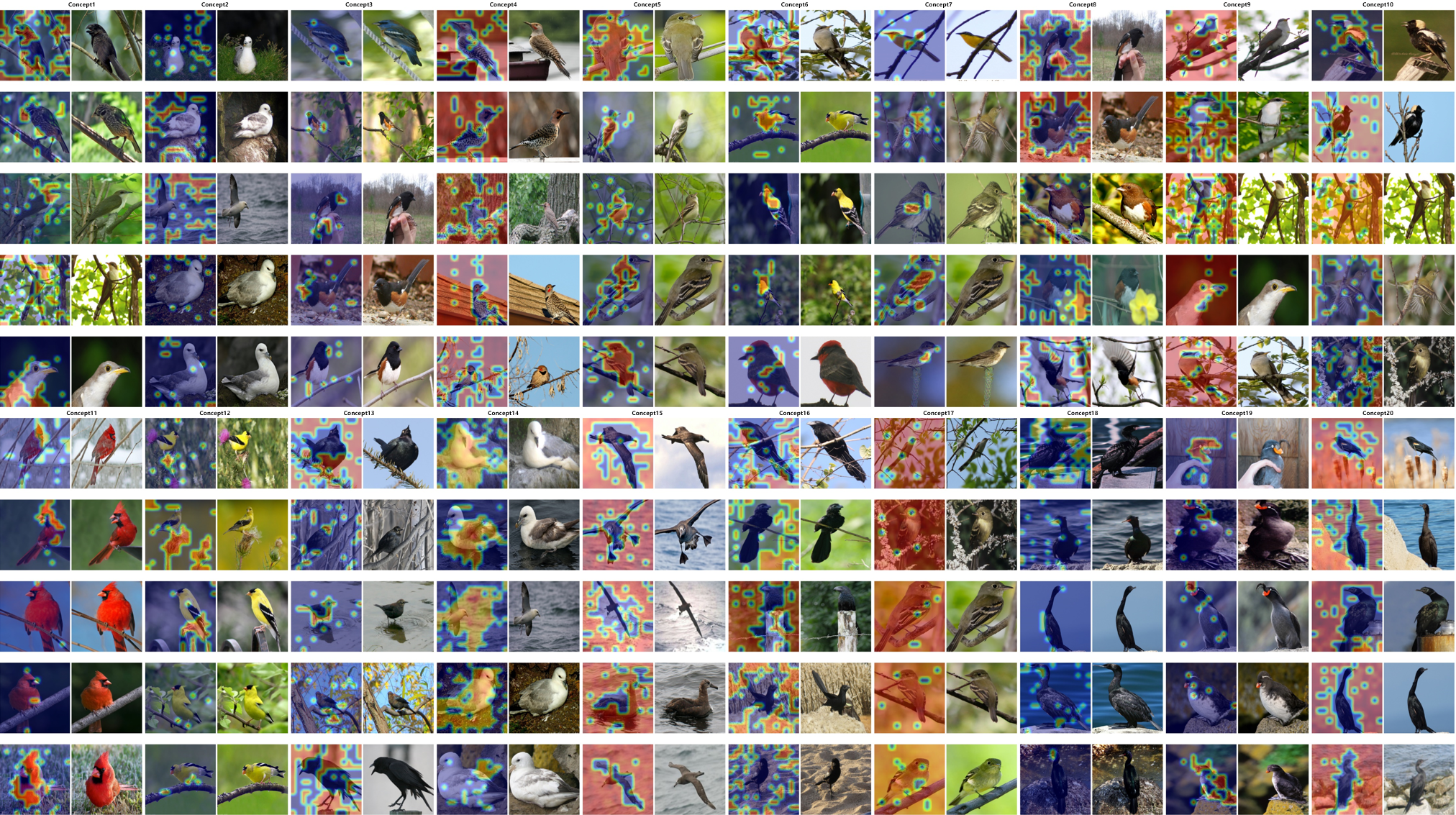}
    \caption{%
    From left to right, the image displays Concepts 1 through 20 as identified in the CUB-200-2011 dataset following ~\cite{wang2023learning}. Each pair of images consists of the masked version ~(with attention activation mask) on the left and the original image on the right. For better visual representation, see \ref{fig:cub_app_first_20} and \ref{fig:cub_app_second_20}. }
    \label{fig:cub_app_entire}
\end{figure*}

\begin{figure*}[t!]
    \centering
    \includegraphics[width=\textwidth]{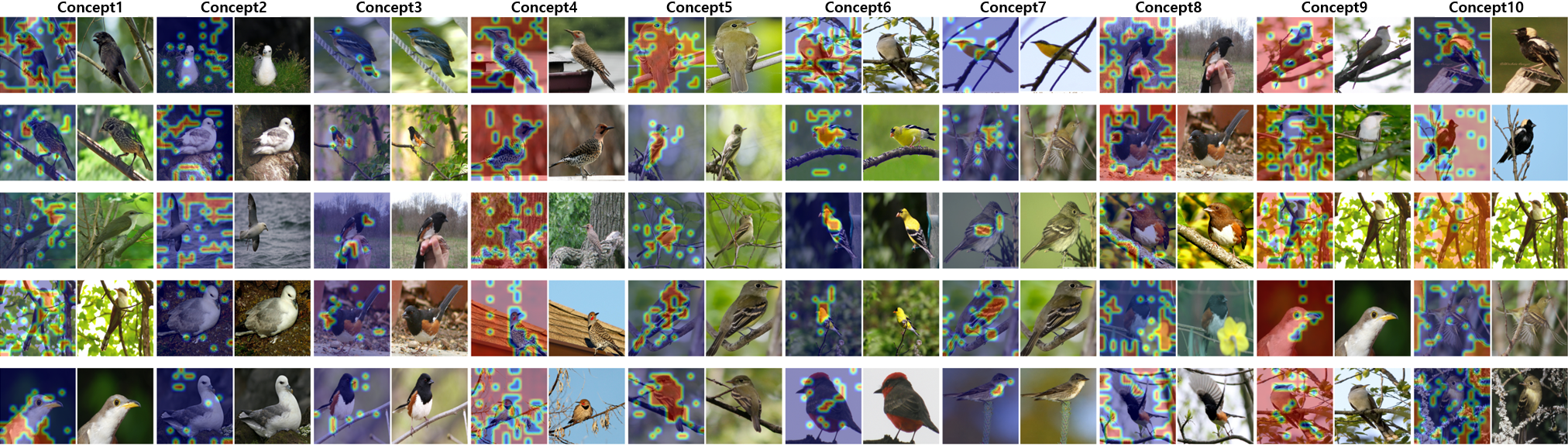}
    \caption{First 10 concepts~(1-10) in CUB-200-2011}
    \label{fig:cub_app_first_20}
\end{figure*}

\begin{figure*}[t!]
    \centering
    \includegraphics[width=\textwidth]{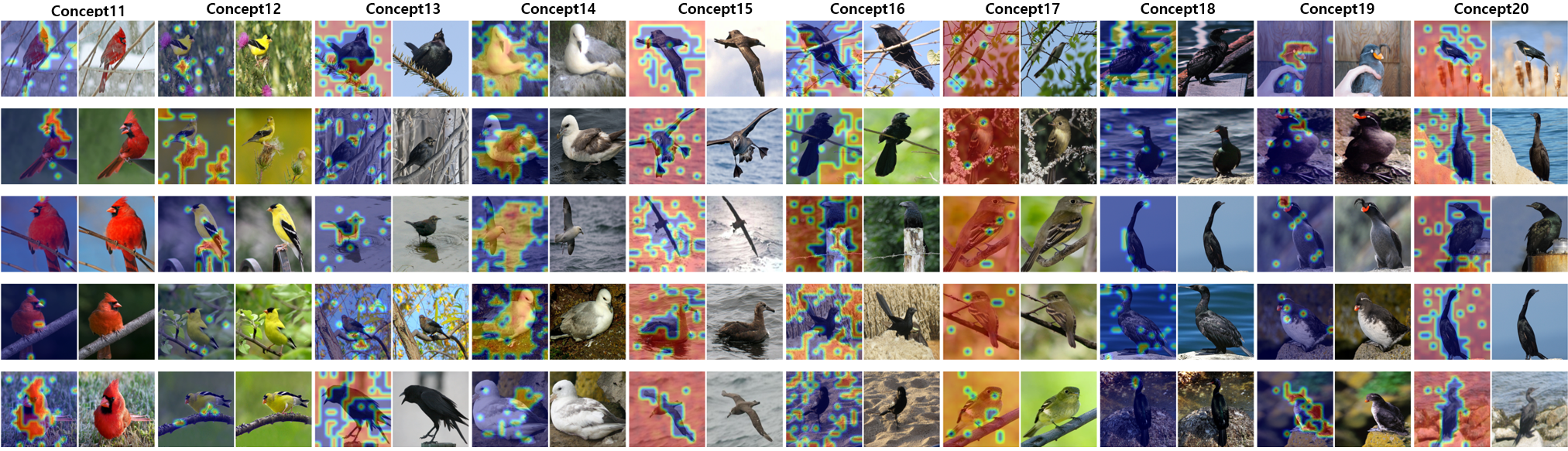}
    \caption{Second 10 concepts~(11-20) in CUB-200-2011}
    \label{fig:cub_app_second_20}
\end{figure*}

\paragraph{Conceptual Visualization} 
For an in-depth look into the learnt concepts within CUB-200-2011 dataset, we adapted our CCT model to 50 distinct classes and 20 latent concepts following ~\cite{wang2023learning}. A comprehensive visualization of these refined latent concepts can be viewed in Figure \ref{fig:cub_app_entire}, \ref{fig:cub_app_first_20} and \ref{fig:cub_app_second_20}. Please note that we present these visualization results without providing concept explanations.

In the CUB-200-2011 dataset, our model's first concept highlights key features of birds such as their head and beak area. Concept 2 focuses on the contours of seagulls by emphasizing the background. Concept 3 considers multiple features like the beak, eyes, and tail, while Concept 4 isolates birds from complex backgrounds. Similarly, Concept 5 outlines the bird's entire body. Concept 6 specializes in highlighting the body area of yellow-bodied birds, and Concept 7 zeroes in on the beak and upper torso. Concept 8 not only highlights contours but also focuses on the bird's eye area. Concepts 9 and 10 share similarities with Concept 8, emphasizing the eye region. Concept 11 stands out by focusing on the head area of red parrots, which is also a feature that catches human attention. Concepts 12 and 13 are dedicated to the bird's feet and lower body areas, respectively. Concept 14 traces the entire body of a seagull, whereas Concept 15 captures the contours of birds in flying poses. Concept 16, on the other hand, focuses on the contours of birds in sitting poses. Concept 17 centers on the eye and mid-body regions, while Concept 18 partially captures key features of long-necked birds. Concept 19 captures unique head and feather structures. Finally, Concept 20 perfectly outlines the bird's contour.

These results demonstrate the model's ability to focus on a wide range of features, from beaks and eyes to tails and feet, reinforcing its classification skills. This is particularly remarkable in an environment where explicit concept explanations are unavailable, highlighting another dimension where our model excels over others.

\paragraph{Image Retrieval and Clustering.}
In addition to its robust classification capabilities, our CCT model excels in the realm of image retrieval. Remarkably, the model achieves this without requiring explicit explanations for the 20 latent concepts it identifies within the CUB-200-2011 dataset. This intuitive clustering of images based on inherent features—ranging from the contours of seagulls to the unique features of a red parrot's head—demonstrates another dimension in which our model surpasses others in the field. Essentially, it clusters semantically related images based on these latent concepts, offering coherent results even in an environment where we don't have access to concept explanation. For instance, images featuring parrots are clustered together, driven by Concept 11's focus on the head area of red parrots. Similarly, images of seagulls are grouped together, guided by Concept 2's emphasis on seagull contours.


\begin{figure}[t!]
    \centering
    \includegraphics[width=0.47\textwidth]{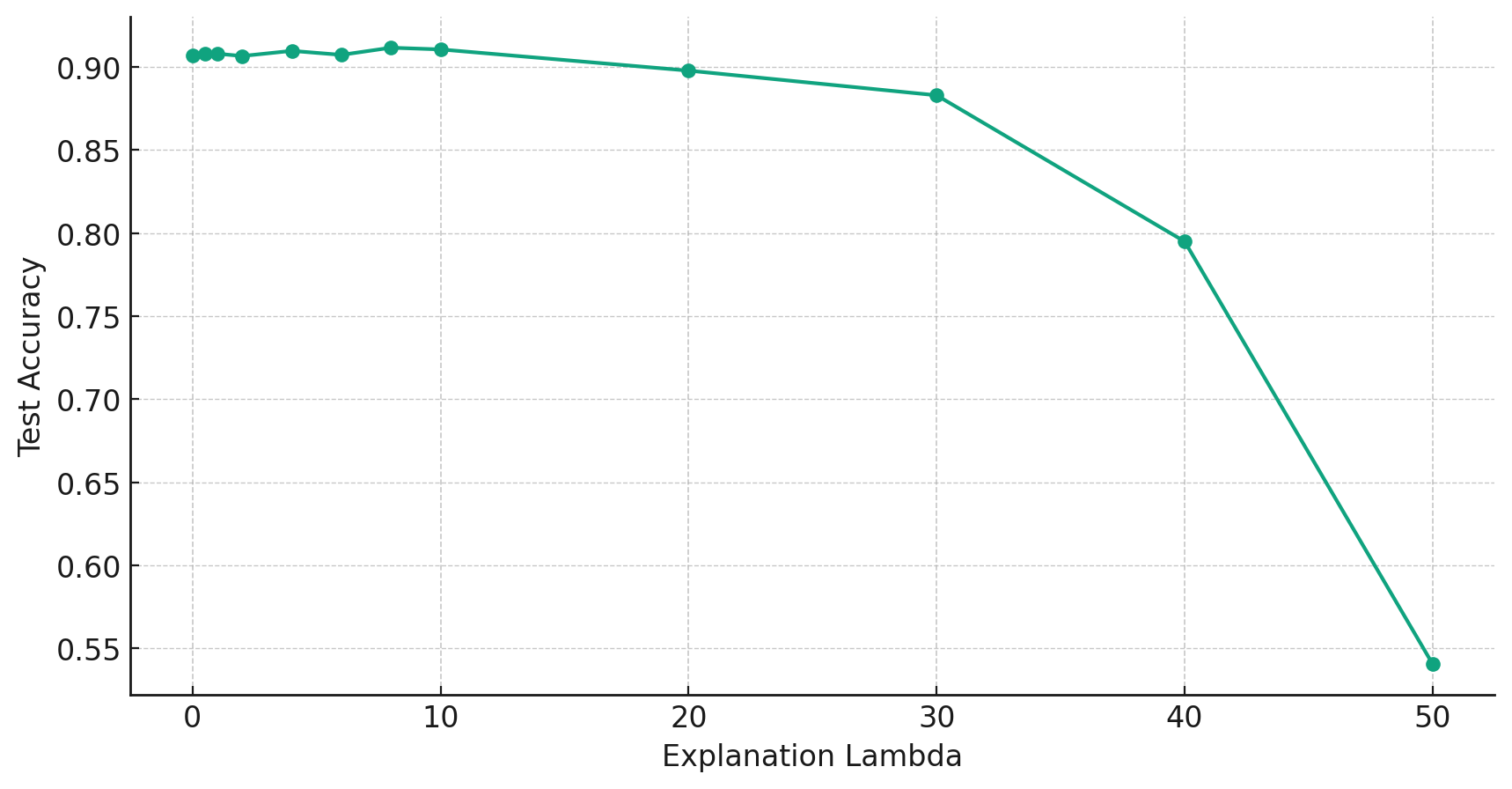}
    \caption{Performance comparison of CCTs on CUB-200-2011 with different value of explanation lambda $\lambda_{expl}$.}
    \label{fig:ablation_expl_lamba}
\end{figure}

\paragraph{Ablation Study.}
In Figure \ref{fig:ablation_expl_lamba}, we can infer that the CCT's performance is sensitive to the explanation lambda hyperparameter $\lambda_{expl}$~(\ref{subsec:training} in the main text). There seems to be an optimal range for explanation lambda, somewhere between 0 and 10, within which the model performs best. Values of $\lambda_{expl}$ higher than 10 lead to progressively worse performance, with severe degradation observed at the highest values of lambda tested.

\begin{figure}[t!]
    \centering
    \includegraphics[width=0.47\textwidth]{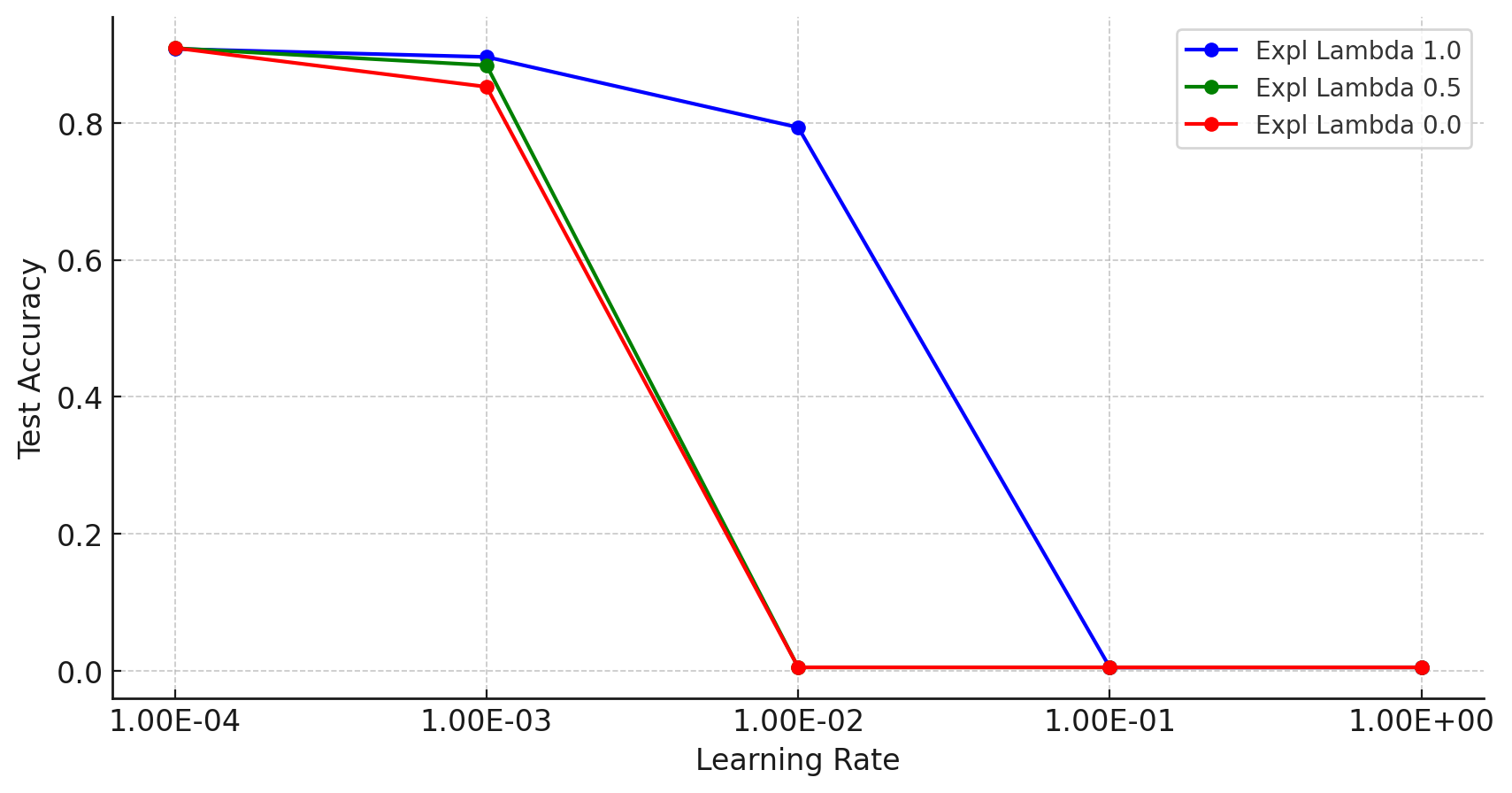}
    \caption{Performance comparison of CCTs on CUB-200-2011 with different value of learning rates under different explanation lambda values.}
    \label{fig:ablation_LR}
\end{figure}

In Figure \ref{fig:ablation_LR}, we can infer that a lower learning rate of 1.00e-4 provides the best model performance, especially when the explanation lambda $\lambda_{expl}$ is set to 0.0. As the learning rate increases, the model's accuracy deteriorates significantly. The extremely low accuracy at higher learning rates (1.00e-01 and 1.00e+00) suggests that the model is unable to learn effectively, likely overshooting the optimal values during training due to too large updates to the model's parameters.

\subsection{ImageNet Experiments}
\label{app_sub:imagenet}

\paragraph{Experiment Design.}
To further confirm CCT's capability to derive latent concepts autonomously, we conducted an experiment on the ImageNet dataset, adhering to the methodology described in ~\cite{wang2023learning}. We chose ViT-S as our backbone architecture, avoiding models with more than 45M parameters, such as ResNet-101, Swin-S, and ConvNeXt-S.

\begin{figure*}[t!]
    \centering
    \includegraphics[width=\textwidth]{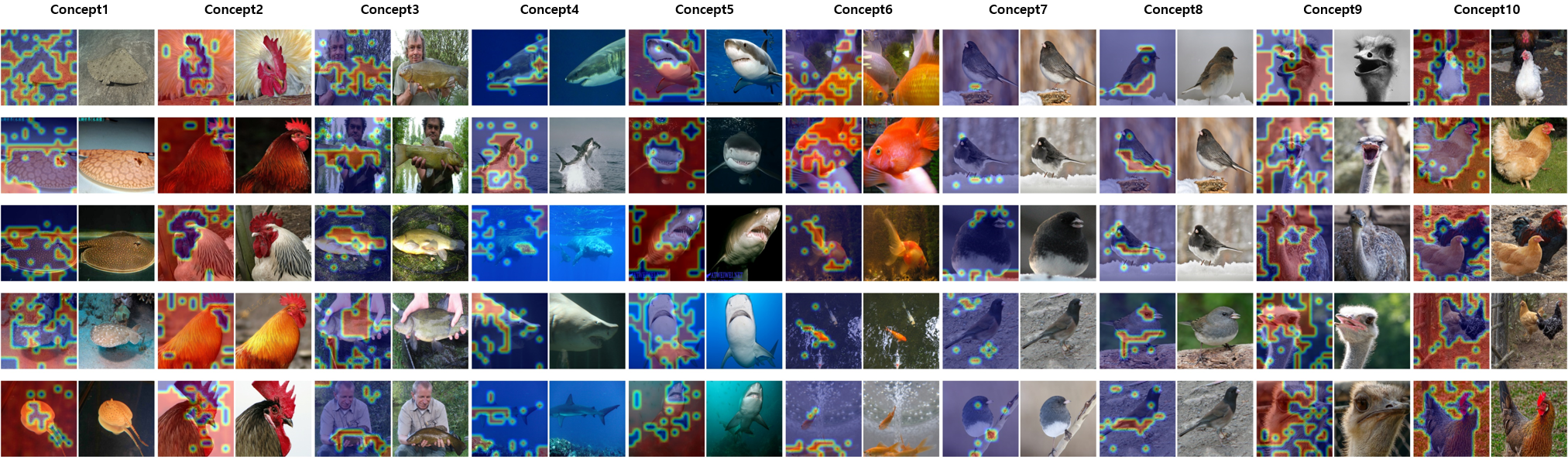}
    \caption{All 10 concepts~(1-10) in ImageNet dataset following ~\cite{wang2023learning}}
    \label{fig:imagenet_10}
\end{figure*}

\paragraph{Conceptual Visualization.}
For an in-depth look into the learnt concepts, we tailored our CCT model to 20 classes and 10 latent concepts, as in ~\cite{wang2023learning}. Figure ~\ref{fig:imagenet_10} provides a detailed visualization of these latent concepts. Please note that we present these visualizaton results without providing concept explanations.

The first concept our model identifies focuses on jelly shellfish, specifically isolating the contours of these creatures against their natural backgrounds. The second concept turns its attention towards chickens, particularly emphasizing the distinct red comb atop their heads. Following this, the third concept excels in separating the various components of fish species, effectively delineating between different parts such as fins, scales, and tails. Moving to marine life, the fourth concept aims to concentrate on the facial attributes of dolphins and sharks, with a particular focus on their mouths. The fifth concept takes this a step further by specifically highlighting the regions around the oral cavities of sharks, setting them apart from other parts of the creature. In a similar aquatic vein, the sixth concept outlines the unique shapes and contours of goldfish, capturing their form effectively. The seventh concept diverges by focusing solely on the feet of birds, whether they are perched or in flight. This is complemented by the eighth concept, which takes a broader approach to birds by concentrating on their ventral regions, capturing details such as feathers and underbellies. The ninth concept specializes in ostriches, particularly focusing on the distinct features that make up their head region. Finally, the tenth concept zeroes in on hens. It particularly focuses on isolating their contours against a variety of backgrounds, thereby allowing for a clearer understanding of the hen's form and structure.

These visualizations demonstrate CCT’s unparalleled ability to focus on semantically meaningful aspects of the images.
\paragraph{Image Retrieval and Clustering.}
In addition to identifying intricate latent concepts, our CCT model demonstrates exceptional capabilities in image retrieval tasks. As illustrated in Fig. ~\ref{fig:imagenet_10}, the model is adept at clustering images that share semantic similarities, thereby reinforcing its utility in generating semantically coherent results. Notably, our CCT achieves this level of clustering without any need for explicit concept explanations. This sets it apart from other models in the field, as it can intuitively group images based on the inherent features recognized through the latent concepts, ranging from the contours of jelly shellfish to the unique features of an ostrich's head. This ability to cluster semantically related images without detailed conceptual guidance emphasizes another aspect where our model excels over others in the domain.

\paragraph{Hyperparameter Settings.}
We follow the official implementation of CT and produce the experimental results, including Table~\ref{tab:eval_imagenet} in the main text. 

The only difference in hyperparameter settings between our CCT, CT, and the vanilla ViT-S is the learning rate for the AdamW optimizer. For the vanilla ViT-S, the learning rate is 0.0001. For CT, the learning rate is 5e-5, following that in their CUB-200-2011 experiments. For our CCT, the learning rate is 0.0001. Table~\ref{tab:hparam_cct_imagenet} shows the shared hyperparameters for the vanilla ViT-S, CT, and all configurations of our CCT. Please refer to the Table \ref{tab:hparam_Proto_archi} for hyperparameter settings used in our baseline experiments.

\paragraph{Limitations of Prototype-based Methods.} 
The observation from testing prototype-based models such as ProtoPFormer, ProtoPool, ProtoPNet, and Deform-ProtoPNet on ImageNet yielded some interesting results. While we initially reported outcomes for only the first 200 labels of ImageNet in Table~\ref{tab:eval_imagenet}, in line with the methodology from ~\cite{wang2023learning}, it was necessary for us to reduce the number of prototypes per instance—a critical hyperparameter for prototype based model—due to computational constraints, whereas our transformer-based model, CCT, managed the task without issue. We hypothesize that this limitation arises from the inherent design of prototype-based models, which utilize 'prototypes' to offer interpretability to the decision-making processes of complex machine learning models in areas such as image classification, object detection, or segmentation. It can be a trivial issue when the size of dataset is small, however, as the complexity of the dataset increases, such as with ImageNet, which contains millions of images across thousands of categories, the number of prototypes required to effectively represent all classes grows. This growth demands significantly more memory and computational resources, particularly GPU RAM, to store and process these prototypes during both training and inference.

\paragraph{Additional Experimental Results.}
\begin{figure}[t!]
    \centering
    \includegraphics[width=0.47\textwidth]{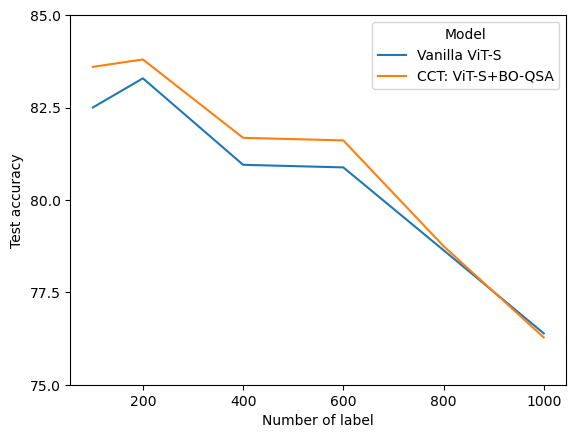}
    \caption{Performance comparison on ImageNet with different number of labels.}
    \label{fig:imagenet_num-label}
\end{figure}
\begin{figure}[t!]
    \centering
    \includegraphics[width=0.47\textwidth]{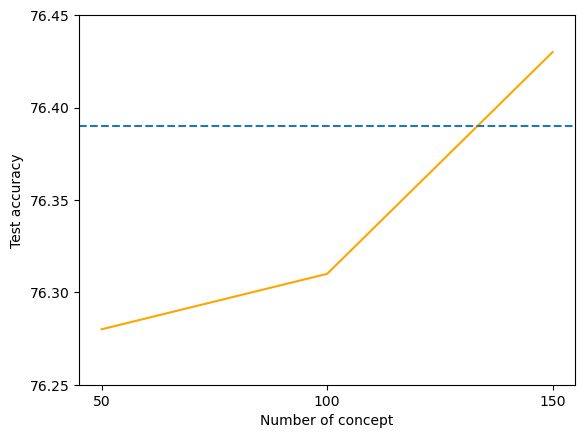}
    \caption{Performance comparison of CCTs on ImageNet with different number of latent concepts. The dashed blue line indicates the test accuracy of vanilla ViT-S.}
    \label{fig:imagenet_num-concept}
\end{figure}
Fig.~\ref{fig:imagenet_num-label} shows the experimental results on ImageNet with different number of labels to train vanilla ViT-S and CCT. We tested both models with the same hyperparameter setting in the main text, and selected the best configuration of CCT, using BO-QSA. Even though the number of latent concepts in CCT is 50 following~\cite{wang2023learning}, CCT mostly outperformed the pretrained backbone. When the number of labels is enormous~$(\geq 800)$, the performance of CCT is similar to or worse than that of the backbone. This is typical because the number of 50 latent concepts we set is insufficient to process all labels.

Therefore, we performed an additional experiment on ImageNet using a total of 1000 classes and different numbers of latent concepts for CCT. 
Fig.~\ref{fig:imagenet_num-concept} demonstrates the performance comparison of CCTs depending on the number of latent concepts. When the number of concepts is set to 150, CCT outperformed the vanilla ViT-S~(the dashed blue line), indicating the contribution of setting the number of latent concepts to the performance of CCT.

\section{Limitations}
\label{app:limit}

In this section, we explain the limitations of our proposed approach. 

First, because of the CCT's architectural characteristics in Section~\ref{sec:method} in the main text, the proposed approach enforces additive contributions from the user-customized concepts to the classification probabilities. In other words, we ignore the latent higher-order relations among the concepts. For instance, in our CUB-200-2011 experiments, we assumed that global and spatial concepts could be represented and learned parallelly in our framework and that there is no correlation between the global concept and the spatial concept, which is independent. However, this is limited because more complex relationships, such as hierarchical properties, can exist among them. It might be addressed by introducing refined architectural properties in our CCT. For example, the Bi-directional Recurrent Unit can be introduced to allow global and spatial concepts to learn each other's conceptual influences. 

Secondly, although, in the experiments of CUB-200-2011 and ImageNet without using explanations, we demonstrate that our CCT can visualize the learned semantic concepts,  we acknowledge that our CCT might be integrated with more sophisticated losses to achieve better visualization of the learned concepts. For example, \cite{yanglearning} leverages losses, including reconstruction, contrastive losses, and various regularizers, to enforce the individual consistency and mutual distinctness of concepts. In our future research, we might use those losses and regularizers to allow our model to achieve better visualization capabilities. 

\end{document}